\documentclass{article}
\usepackage{ijcai24}

\usepackage{times}
\usepackage{soul}
\usepackage{url}
\usepackage[hidelinks]{hyperref}
\usepackage[utf8]{inputenc}
\usepackage[small]{caption}
\usepackage{graphicx}
\usepackage{amsmath}
\usepackage{amsthm}
\usepackage{booktabs}
\usepackage{algorithm}
\usepackage{algorithmic}
\usepackage[switch]{lineno}

\usepackage{subcaption}
\usepackage{multirow}
\usepackage{amssymb}

\title{Wearable Sensor-Based Few-Shot Continual Learning on Hand Gestures for Motor-Impaired Individuals via Latent Embedding Exploitation}

\author{
Riyad Bin Rafiq$^1$
\and
Weishi Shi$^1$\And
Mark V. Albert$^{1,2}$
\affiliations
$^1$Department of Computer Science and Engineering, University of North Texas\\
$^2$Department of Biomedical Engineering, University of North Texas\\
\emails
riyadbinrafiq@my.unt.edu,
\{weishi.shi, mark.albert\}@unt.edu
}

\begin{document}

\maketitle
\begin{abstract}
Hand gestures can provide a natural means of human-computer interaction and enable people who cannot speak to communicate efficiently. Existing hand gesture recognition methods heavily depend on pre-defined gestures, however, motor-impaired individuals require new gestures tailored to each individual's gesture motion and style. Gesture samples collected from different persons have distribution shifts due to their health conditions, the severity of the disability, motion patterns of the arms, etc. In this paper, we introduce the Latent Embedding Exploitation (LEE) mechanism in our replay-based Few-Shot Continual Learning (FSCL) framework that significantly improves the performance of fine-tuning a model for out-of-distribution data. Our method produces a diversified latent feature space by leveraging a preserved latent embedding known as \textit{gesture prior knowledge}, along with \textit{intra-gesture divergence} derived from two additional embeddings. Thus, the model can capture latent statistical structure in highly variable gestures with limited samples. We conduct an experimental evaluation using the SmartWatch Gesture and the Motion Gesture datasets. The proposed method results in an average test accuracy of 57.0\%, 64.6\%, and 69.3\% by using one, three, and five samples for six different gestures. Our method helps motor-impaired persons leverage wearable devices, and their unique styles of movement can be learned and applied in human-computer interaction and social communication. Code is available at: \url{https://github.com/riyadRafiq/wearable-latent-embedding-exploitation}.
\end{abstract}

\section{Introduction}
Hand gestures are a flexible and intuitive means of communication for human beings. With the advancement of wearable sensors and machine learning, gesture recognition has become quite popular for communication, smart home appliances, interactive entertainment, etc.~\cite{rafiq2023lstm,guo2021human}. Gesture-based interactions with wearables depend on specific presumptions about users' motor abilities. As a consequence, people with motor impairments face challenges in performing gestures with wearables that are widely adopted for the general public~\cite{siean2021wearable}. The severity of motor impairments leads to a different pattern of motion gestures and creates individual differences among the users~\cite{vatavu2022understanding}. It is social discrimination for this underrepresented population as it deprives them of completely leveraging those wearable devices. As the United Nations Sustainable Development Principle is \textit{Leave no one behind}, increasing independence and including people with disabilities aid in achieving UN Sustainable Development Goals \textit{Good health and well-being and Reduce inequalities}~\cite{yu2023quality}.

\begin{figure}[t!]
     \centering
     \begin{subfigure}[b]{0.18\textwidth}
         \centering
         \includegraphics[width=\textwidth]{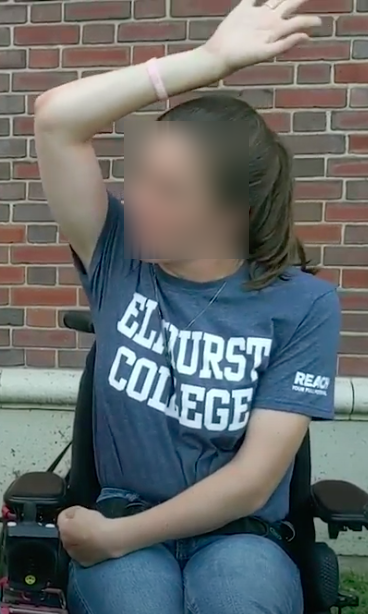}
         \caption{}
         \label{fig1a}
     \end{subfigure}
     \hspace{1mm}
     \begin{subfigure}[b]{0.22\textwidth}
         \includegraphics[width=\textwidth]{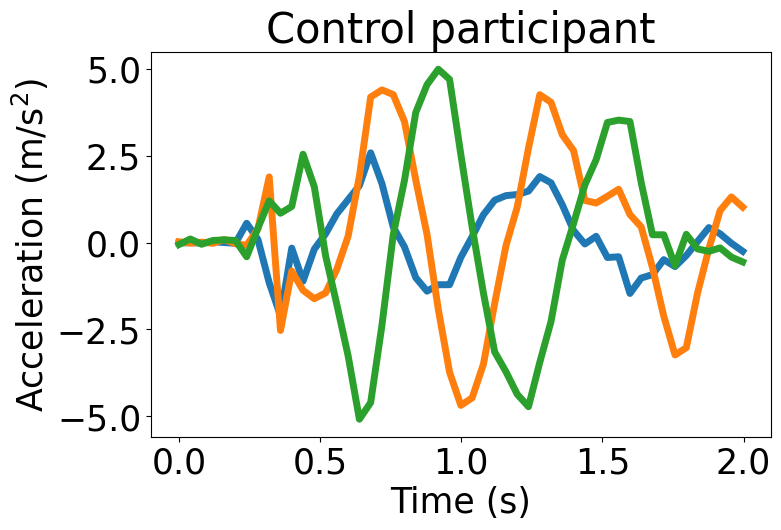}
         \includegraphics[width=\textwidth]{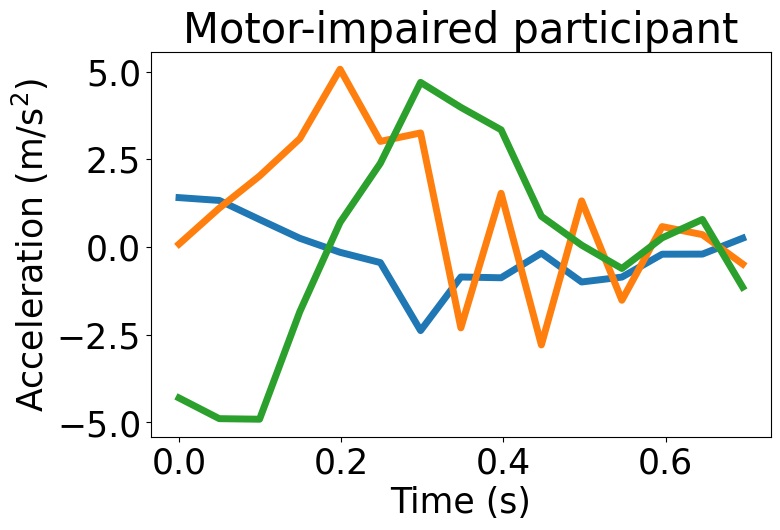}
         \caption{}
         \label{fig1b}
     \end{subfigure}
    \caption{(a) An individual lacking fine motor skills performs hand gestures. (b) Sensor-based gesture samples of two different participants including a control participant (top) and a motor-impaired participant (bottom). Data samples are more variable and noisy for a motor-impaired individual rather than a control participant. Blue, orange, and green lines are acceleration values along the x, y, and z-axis respectively.}
    \label{fig1}
    \vspace{-4mm}
\end{figure}

To tackle the problem, a large-scale labeled dataset is expected to build a robust hand gesture recognition method. However, this is impractical and cumbersome for motor-impaired individuals to participate in vast data collection. The transfer learning approach has been used to solve the problem. In transfer learning, a model is trained on a source domain and then fine-tuned to a target domain by transferring knowledge from the prior learned task~\cite{zhuang2020comprehensive}. But fine-tuning shows worse performance in a target domain with out-of-distribution samples~\cite{kumar2022fine}. Therefore, applying transfer learning alone cannot solve the problem, as the gesture data from the motor-impaired population are more variable and noisy than the control population (Figure~\ref{fig1}), and limited data samples might not help the deep learning model capture the diverse patterns among each individual. To utilize limited training data, a unique approach has been proposed~\cite{finn2017model} and in our case, few-shot transfer learning is an applicable solution.

In our case, another real-world problem is that all the gesture classes may not be available initially. For example, standard pre-defined gestures can be difficult to perform for individuals lacking fine motor skills. New unseen gestures may become accessible incrementally if motor-impaired individuals want to input their flexible and custom gestures. This context is referred to as a continual learning setting as the model involves learning a disjoint set of classes incrementally~\cite{parisi2019continual}. Continual learning posts two challenges, namely catastrophic forgetting~\cite{french1999catastrophic} when the model's performance drops drastically on old classes and overfitting when the model is not capable of learning generalized features with a few training examples~\cite{gidaris2018dynamic}.     

Many continual learning approaches including parameter regularization, functional regularization, replay strategy, etc. have become popular at present~\cite{van2019three}. In this paper, we propose a novel method called Latent Embedding Exploitation (LEE) in our replay-based few-shot continual learning framework that can learn gesture classes incrementally from motor-impaired people. Specifically, in our framework, we utilize three latent embeddings from the feature extractor of a pre-trained model which is trained on the control subjects' gesture samples. The three embeddings are:
\begin{itemize}
    \item a preserved latent embedding works as \textit{gesture prior knowledge},
    \item two additional latent embeddings known as temporary and learned embedding maintain a \textit{intra-gesture divergence}.
\end{itemize}
They jointly aid a pre-trained model to be fine-tuned effectively with a few training examples from a motor-impaired individual. Ideally, the goal of LEE is to navigate the learned feature space toward a rich and diversified feature representation for variable and noisy data. Thus the fine-tuned model can capture the diverse pattern of unseen gesture classes with a few training examples. As a result, motor-impaired people can take full advantage of wearable devices with our proposed method. The major contributions of this paper are as follows:

\begin{itemize}
    \item We explore wearable sensor-based hand gestures from the underrepresented population. In addition, we introduce the LEE mechanism in our replay-based few-shot continual learning framework that formulates the diverse gesture samples into a heterogeneous feature representation. Hence, the pre-trained model can be fine-tuned competently with a few training samples for each unseen class in a continual learning setup.
    \item We utilize two publicly available gesture datasets to demonstrate the performance of the proposed method. Our proposed method achieves competitive performance compared to existing methods.
    \item We experimentally show how latent embeddings can be leveraged to improve the performance of fine-tuning in the limited data of shifted distribution. 
\end{itemize}

\section{Related Work}
A wide range of hand gesture recognition techniques has been explored by utilizing images and videos~\cite{hu2018learning,zhou2021regional}, electromyography (EMG)~\cite{caramiaux2015understanding} and wearable-sensors~\cite{laput2019sensing,kunwar2022robust}. Among these techniques, vision-based approaches show poor performance due to complex backgrounds, varying light conditions, and the presence of another person in the background~\cite{pisharady2015recent,mohamed2021review}. Moreover, high computational power is required to analyze high-quality video sequences and individuals might not be comfortable sharing live video streams due to privacy. On the contrary, sensors including accelerometers and gyroscopes are low-cost and widely available in current wearable devices. Therefore, in our work, we utilize wearable sensor-based motion data to solve the problem.

Prior work has been done using hand-crafted features such as mean, variance, median, maximum, minimum,  etc. for classifying the hand gestures~\cite{xie2016accelerometer}. However, domain expertise is needed to prepare the necessary hand-crafted features which is time-consuming and the methods that utilize those features show poor performance in practice. On the contrary, deep learning approaches such as Convolutional Neural Networks (CNN), Recurrent Neural Networks (RNN), and Transformer architectures have demonstrated significant performance in hand gesture classification with automatically extracted features from training data examples~\cite{kunwar2022robust,nguyen2021gesture,li2019skeleton}. 

The goal of few-shot continual learning is to train new classes incrementally with few data instances. In order to tackle few-shot learning problem, metric-learning~\cite{kaya2019deep} , meta-learning~\cite{finn2017model} and multi-task learning~\cite{zhang2021survey} have been proposed. In the field of hand gesture recognition, camera data~\cite{wu2012one,stewart2020online} and EMG signals~\cite{rahimian2021few} have been used for few-shot learning. However, Xu et al.~\shortcite{xu2022enabling} proposed a hand gesture customization framework that can learn novel hand gesture classes incrementally with a few training examples. Kimura~\shortcite{kimura2022self} also proposed a self-supervised method for few-shot hand gesture recognition using wearable sensor data. However, they used only control participants' data. In addition, these methods would not work for motor-impaired individuals as the data instances are varying and noisy. Although Malu et al.~\shortcite{malu2018exploring} and Kim et al.~\shortcite{kim2019towards} explored the smartwatch interactions for people with upper body motor impairments, they experimented with the touch gestures only. The primary difference between our approach and existing works is that they did not consider the out-of-distribution samples for motion gestures. Directly fine-tuning a pre-trained model may cause a sudden performance decay. The objective of our method is to adapt a model that generates an enhanced feature representation via \textit{gesture prior knowledge} exploitation and \textit{intra-gesture divergence} exploration to incrementally learn novel gestures with few training examples from motor-impaired individuals.

\section{Proposed Method}
\begin{figure*}[ht!]
    \centering
    \includegraphics[width=0.75\textwidth]{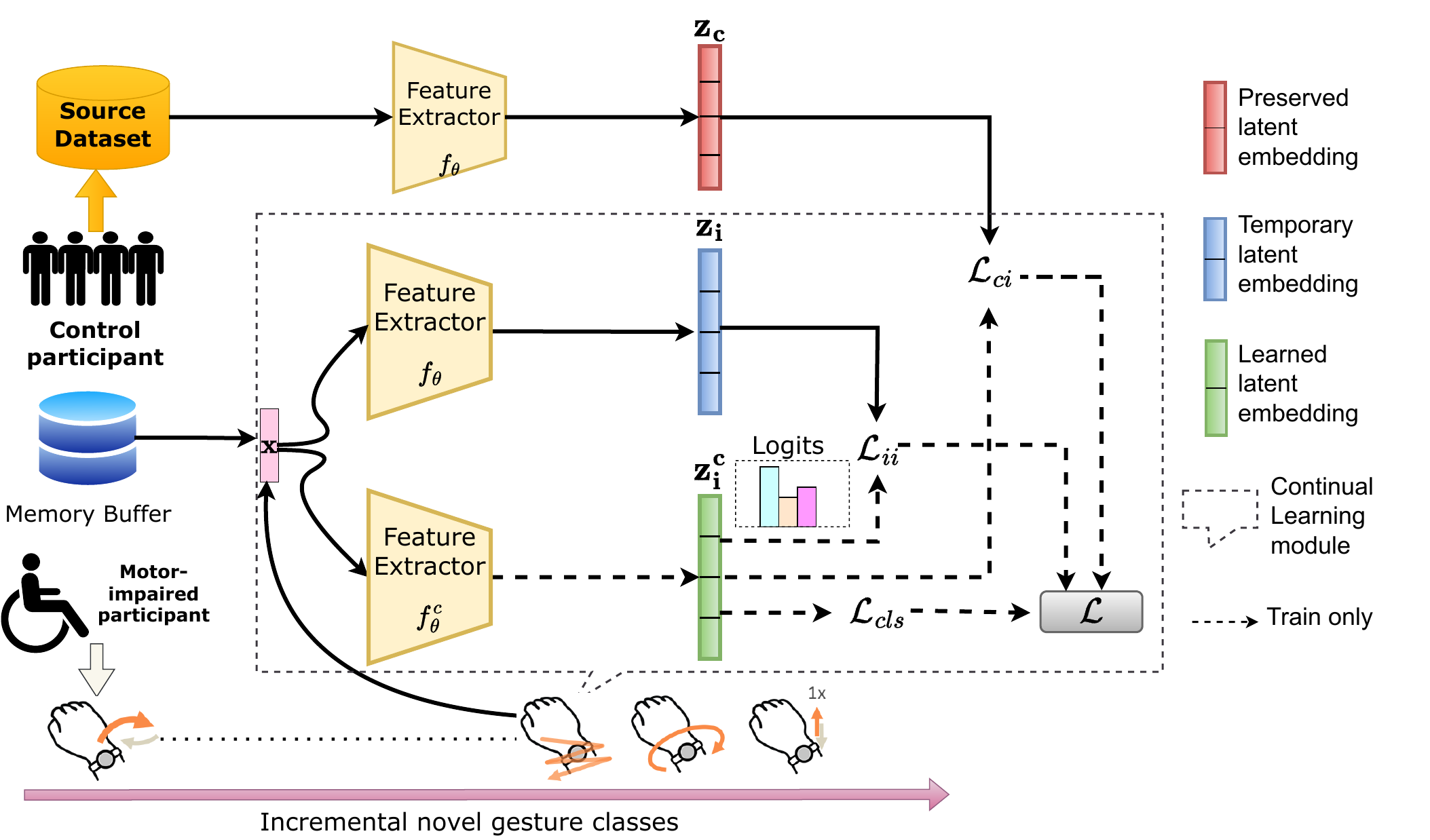} 
    \caption{The complete framework containing the LEE mechanism. A latent embedding, $\mathbf{z_c}$ from the control population is preserved to work as \textit{gesture prior knowledge}. In addition to it, two latent embeddings, $\mathbf{z_i}$ and $\mathbf{z_i^c}$ function to maintain \textit{intra-gesture divergence}. The memory buffer saves the training samples from old gesture classes and provides them while training on a novel class.}   
    \label{fig2}
    \vspace{-4mm}
\end{figure*}

\subsection{Problem Statement}
A domain is defined as a joint probability distribution ${\mathbb{P}_{x,y}}$ on {$\mathcal{X}$} {$\times$} {$\mathcal{Y}$}, where {$\mathcal{X}$} and {$\mathcal{Y}$} denote the instance space and label space, respectively~\cite{ding2017deep,qian2021latent}. In our setting, we have two domains including source domain, {$\mathcal{D}^{s} = \{(x_i, y_i)\}_{i=1} ^ {n_s} $} and target domain, {$\mathcal{D}^{t} = \{(x_i, y_i)\}_{i=1} ^ {n_t}  $} where {${n_t} << {n_s}$}. Each sample, $x \in \mathbb{R}^{L \times 3}$ denotes a signal of $L$ length with three-axis motion values collected from wearable sensors at each timestamp. The two domains have the same feature space ({$\mathcal{X}^{s} = \mathcal{X}^{t}$}) but different label spaces ({$\mathcal{Y}^{s}$} {$\neq$} {$\mathcal{Y}^{t}$}). In addition to it, they have different probability distributions i.e. {$P^{s}(x_i, y_i)$} {$\neq$} {$P^{t}(x_i, y_i)$}. The data distribution for $\mathcal{D}^{t}$ is harder to learn than $\mathcal{D}^{s}$ due to high-level noise  e.g. $\text{H}[P^t(x)] >> \text{H}[P^s(x)]$ where \text{H} denotes the entropy. Therefore, any statistical learner needs more samples from $\mathcal{D}^{t}$ to converge to achieve the same level of accuracy as models trained on $\mathcal{D}^{s}$. In {$\mathcal{D}^{t}$}, the classes {$\mathcal{C}_{1}$}, {$\mathcal{C}_{2}$}, ..., {$\mathcal{C}_{n}$} are incrementally accessible at the training time. While accessing a new class, $\mathcal{C}_i$, a memory buffer stores training examples from old classes such as {$\mathcal{C}_{1}$}, {$\mathcal{C}_{2}$}, ....., {$\mathcal{C}_{i-1}$}. Motor-impaired individuals may face challenges in providing many consistent data samples. As a result, it becomes more difficult to train the model with such limited samples. The goal is to build a pre-trained model in the source domain {$f: {\mathcal{X}^{s}} \rightarrow {\mathcal{Y}^{s}} $} that can be fine-tuned to learn currently available classes from the target domain with few training examples. We design the method in such a way that should reduce the memory usage compared to traditional replay buffers.
    
\subsection{Overall Framework}
Sensor readings from motor-impaired individuals vary substantially due to factors such as health conditions, severity of disability, movement patterns of arms, etc. (Figure~\ref{fig1b}). Therefore, it is critical for a pre-trained model to adapt to such data samples from sensor readings. Machine learning models without considering out-of-distribution data often result in large performance degradation. For example, a model trained on the data from control participants often fails to capture unique patterns and performs poorly on specific populations such as Parkinson's patients~\cite{bin2020validation}. It is difficult to collect a large volume of diverse labeled data from motor-impaired individuals. Moreover, individuals may need their own, custom flexible gestures from time to time in human-computer interaction and social communication. 

In this paper, we propose a novel technique where the model learns new gesture classes incrementally by utilizing multifunctional latent embeddings. Our method considers three latent embeddings instead of a single representation compared to Autoencoders~\cite{bank2020autoencoders}. As shown in Figure~\ref{fig2}, a latent embedding from the control population is preserved by leveraging the feature extractor (deep encoder) of a pre-trained model. This preserved latent embedding works as \textit{gesture prior knowledge} to assist the model to incrementally learn unseen out-of-distribution gesture samples and prevent overfitting. Two additional identical feature extractors are utilized to produce two latent embeddings with available gesture classes from the motor-impaired subjects, and one of them is being updated during training. As a consequence, the learned latent embedding has a strong capability in classifying highly variable data during inference. The total loss of the model in weighted summation is as follows: 
\vspace{-1mm}
\begin{equation}
    \mathcal{L} = \alpha\mathcal{L}_{ci} + \beta\mathcal{L}_{ii} + \mathcal{L}_{cls} \
\end{equation}

where {$\mathcal{L}_{ci}$} is the loss of discrimination between preserved and learned embedding, {$\mathcal{L}_{ii}$} is the loss of discrimination between temporary and learned embedding and {$\mathcal{L}_{cls}$} is the classification loss. $\alpha$ and $\beta$ are trade-off hyper-parameters where $\alpha+\beta=1$. While minimizing the loss in the training stage, the model exhibits complementary learning behavior by adaptively adjusting its focus between exploitative and explorative representation learning. 

\subsection{Complementary Learning Paradigm}
The complementary learning system plays an important role in the human brain where the hippocampus and the neocortex function in a complementary manner to learn complex behavior~\cite{perrusquia2022human,blakeman2020complementary}. Our learning strategy for new classes is inspired by this \textit{complementary learning paradigm}. The representation space holds the same classes closer under the effect of classification objective while training a deep learning model. However, the model struggles to learn a robust latent space with fewer training examples from out-of-distribution data~\cite{yang2021free}. Therefore, we model the representation space generation process by utilizing both the classification and the embedding discrimination objectives. In our method, we introduce three multipurpose latent embeddings including preserved control embedding, temporary sample embedding, and learned embedding. The preserved control embedding contains the representation space of the expected pattern of gestures from the source domain. The learned embedding constructs a latent statistical structure with the help of preserved and temporary embedding. Thus, the constructed latent space sufficiently narrows the features of the same-class data with limited training examples. 

\subsubsection{Latent Embedding Exploitation}
The learned latent embedding exploits the preserved latent embedding (\textit{gesture prior knowledge}) to enlarge and diversify the feature space. The network architecture is expected to increase the similarity of the feature space between the preserved and the learned embedding. We utilize the feature extractor, {$f_\theta$} (deep encoder), from the pre-trained model to preserve a latent embedding, $\mathbf{z_c}=$ {$f_\theta(\mathcal{X}^{s})$} where $\mathcal{X}^{s}$ is the feature space from the source domain i.e. control participants. The input of our model is denoted as \textbf{x}. In our continual learning module, two additional identical feature extractors,  {$f_\theta$} and {$f_\theta^c$} draw out temporary latent embedding, {$\mathbf{z_i}$}$=f_\theta(\textbf{x})$ and learned latent embedding, {$\mathbf{z_i^c}$}$=f_\theta^c(\textbf{x})$ respectively. To this end, we require to expand the similarity between {$\mathbf{z_c}$} and {$\mathbf{z_i^c}$} as the following loss:  
\vspace{-1mm}
\begin{equation}
    \mathcal{L}_{ci}(f_\theta^c ; \textbf{x}, \mathbf{z_c}) = 1 -  \mathcal{S}_{c}(\mathbf{z_c}, f_\theta^c(\textbf{x}))
\end{equation}

where {$\mathcal{S}_{c}$} is the cosine similarity between $\mathbf{z_c}$ and $f_\theta^c(\textbf{x})$ and it can be defined as follows:
\vspace{-1mm}
\begin{equation}
    \mathcal{S}_{c}(\mathbf{z_c}, \mathbf{z_i^c}) = \frac{\mathbf{z_c} \cdot \mathbf{z_i^c}}{||\mathbf{z_c}|| ||\mathbf{z_i^c}||}
\end{equation} 

\subsubsection{Latent Embedding Exploration}
Simultaneously, the learned latent embedding aims to maximize the distance from the identical temporary sample embedding throughout the training which is called \textit{intra-gesture divergence}. This action works as a way of exploration for a wide feature space. As a result, the learned latent embedding captures tailored and generalizable feature representation to learn novel gesture classes. To minimize the similarity between {$\mathbf{z_i}$} and {$\mathbf{z_i^c}$} is identical as follows:

\begin{equation}
    \mathcal{L}_{ii}(f_\theta, f_\theta^c ; \textbf{x}) = \mathcal{S}_{c}(f_\theta(\textbf{x}), f_\theta^c(\textbf{x}))
\end{equation}

\subsubsection{Learning Objective}
The learning objective of the model is to identify the gesture classes which is a transformation of the input sensor signals to a gesture category. Therefore, we utilize class labels in the final classification layer to guide the learned latent embedding during the training stage. We adopt standard cross-entropy loss for the classification task:
\vspace{-1mm}
\begin{equation}
    \mathcal{L}_{cls}(f_\theta^c ; \mathcal{X}^{t}, \mathcal{Y}^{t}) = - \mathbb{E}_{(\textbf{x},y)\in\mathcal{X}^{t} \times \mathcal{Y}^{t}} \sum_{c=1}^{C} y \log \delta_{c}(f_\theta^c(\textbf{x})) 
\end{equation}

where {$C$} represents the number of classes, {$y$} is the true gesture label, {$\delta_{c}(f_\theta^c(\textbf{x}))$} is the predicted probability and {$\delta_c$} is the softmax function.  

\begin{figure*}[ht!]
    \centering
    \includegraphics[width=13.5cm]{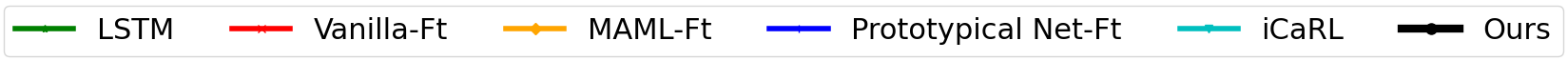}
    \centering
    \begin{minipage}[b]{0.25\textwidth}
        \centering
        \includegraphics[width=\textwidth]{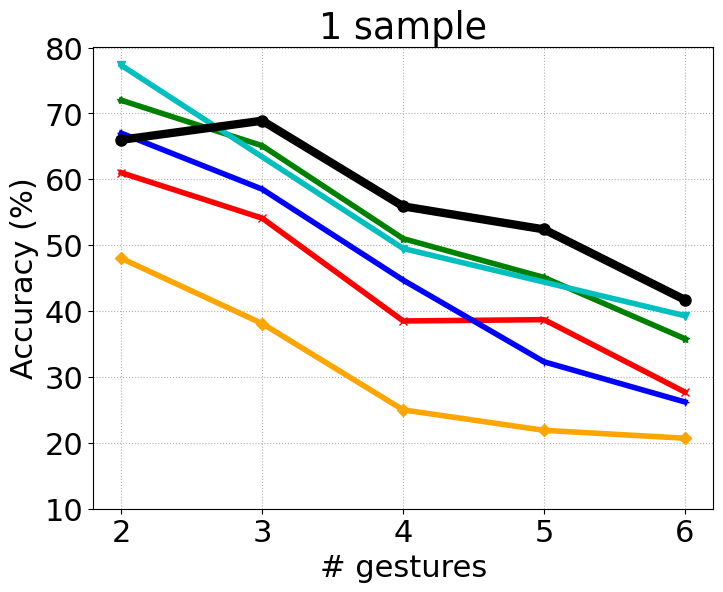}
    \end{minipage}
    \begin{minipage}[b]{0.25\textwidth}
        \centering
        \includegraphics[width=\textwidth]{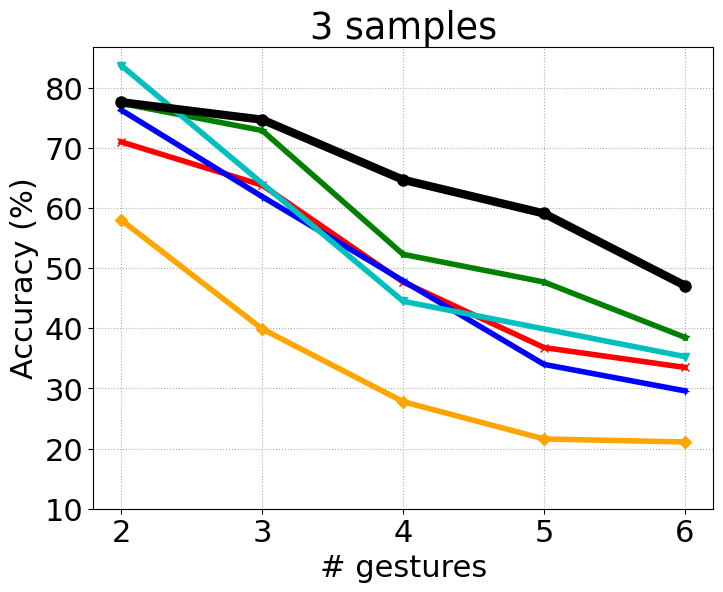}
    \end{minipage}
    \begin{minipage}[b]{0.25\textwidth}
        \centering
        \includegraphics[width=\textwidth]{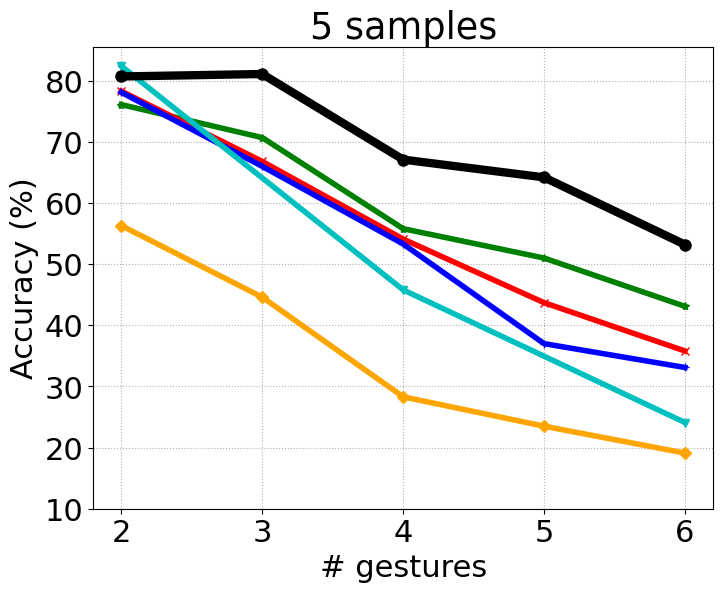}
    \end{minipage}
    \centering
    \begin{minipage}[b]{0.25\textwidth}
        \centering
        \includegraphics[width=\textwidth]{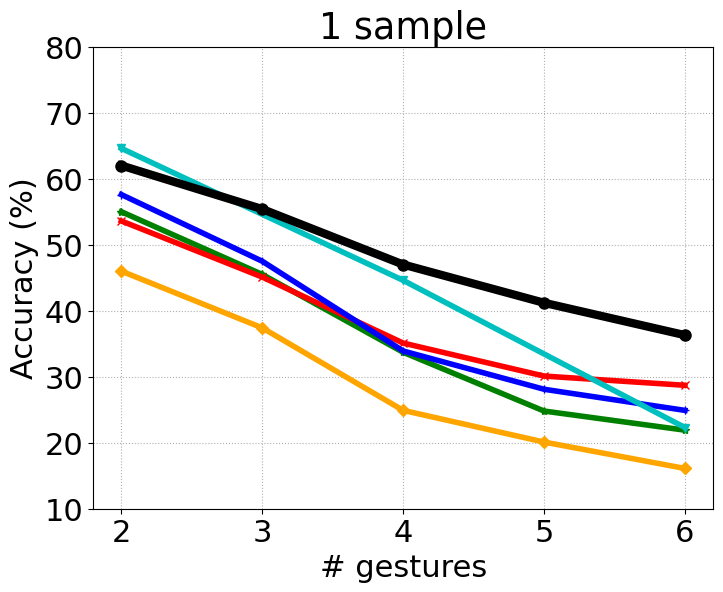}
    \end{minipage}
    \begin{minipage}[b]{0.25\textwidth}
        \centering
        \includegraphics[width=\textwidth]{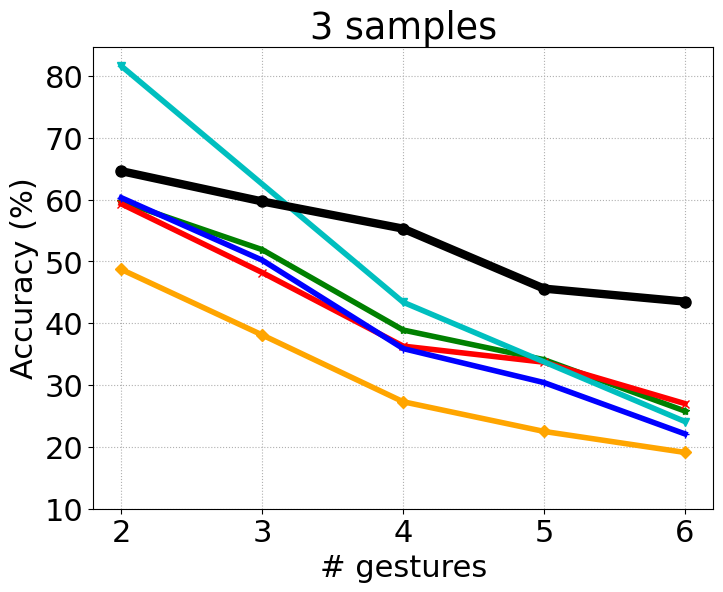}
    \end{minipage}
    \begin{minipage}[b]{0.25\textwidth}
        \centering
        \includegraphics[width=\textwidth]{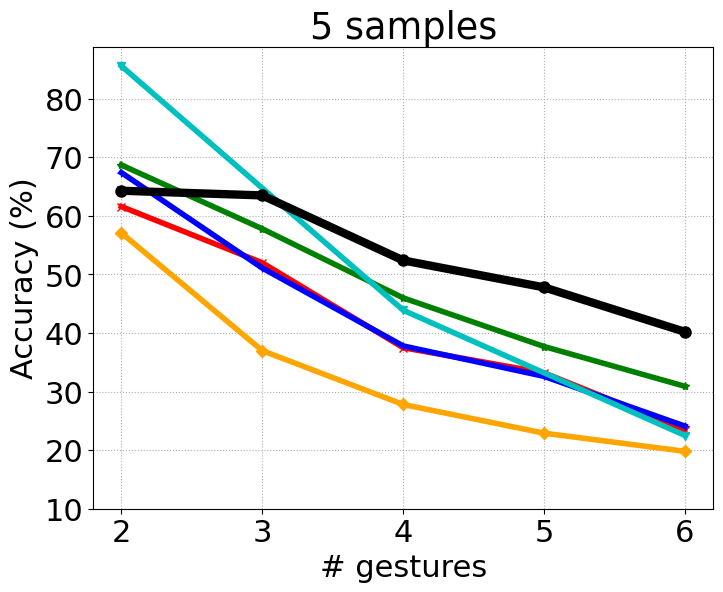}
    \end{minipage}
    \centering
    \begin{minipage}[b]{0.25\textwidth}
        \centering
        \includegraphics[width=\textwidth]{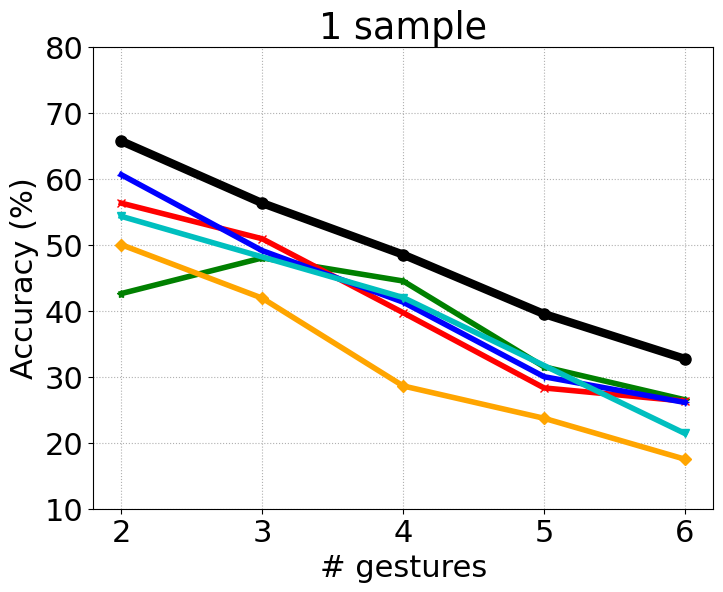}
    \end{minipage}
    \begin{minipage}[b]{0.25\textwidth}
        \centering
        \includegraphics[width=\textwidth]{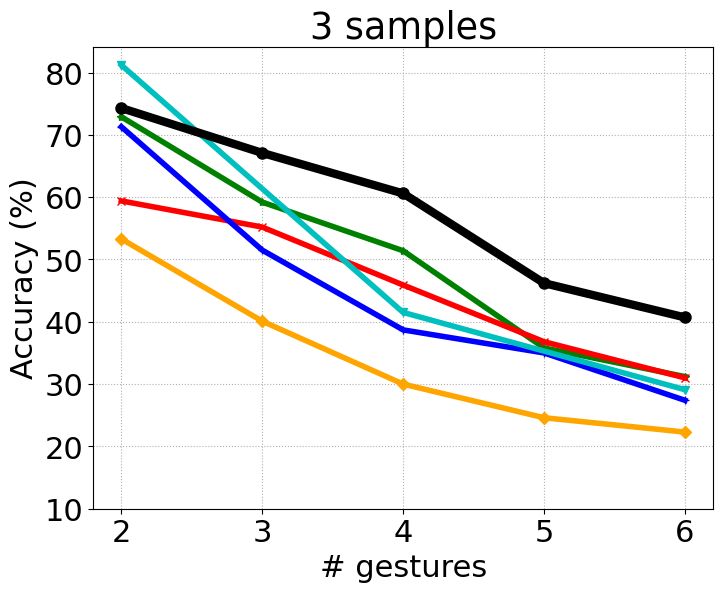}
    \end{minipage}
    \begin{minipage}[b]{0.25\textwidth}
        \centering
        \includegraphics[width=\textwidth]{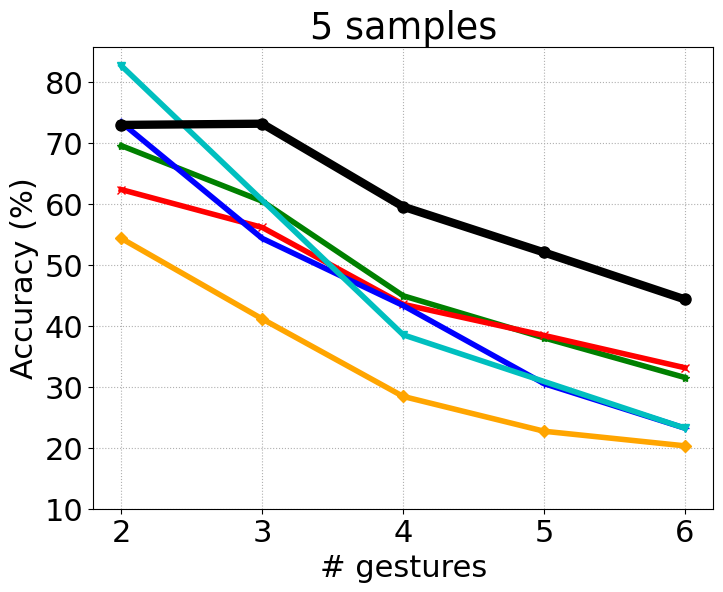}
    \end{minipage}
    
    \caption{Test accuracies for a motor-impaired individual with Spinal cord injury (top row), an individual with Parkinson's disease (middle row), and a participant with Multiple sclerosis (bottom row) in a few-shot continual learning setting. The accuracy represents the total accuracy over all the gesture classes encountered trained with one, three, and five samples.}
    \label{fig3}
    \vspace{-4mm}
\end{figure*}

\begin{table*}[t]
\centering
\small
\begin{tabular}{c|cccccc}
\hline
\multicolumn{1}{c|}{\multirow{2}{*}{Gesture classes}} & \multicolumn{6}{c}{\textbf{Methods}}                                          \\ \cline{2-7} 
\multicolumn{1}{c|}{}                                 & LSTM      & Vanilla-Ft & MAML-Ft   & Prototypical Net-Ft & iCaRL & \textbf{Ours-LEE} \\ \hline
\textit{Gesture 1}                                             & \underline{0.44 $\pm$ 0.08} & 0.31 $\pm$ 0.07  & 0.13 $\pm$ 0.10 & 0.24 $\pm$ 0.09 & 0.01 $\pm$ 0.02 & \textbf{0.53 $\pm$ 0.19} \\
\textit{Gesture 2}                                             & \underline{0.21 $\pm$ 0.11} & 0.16 $\pm$ 0.05  & 0.12 $\pm$ 0.08 & 0.10 $\pm$ 0.04 & 0.04 $\pm$ 0.08 & \textbf{0.28 $\pm$ 0.13} \\
\textit{Gesture 3}                                             &      
0.35 $\pm$ 0.15 & \underline{0.40 $\pm$ 0.07}  & 0.12 $\pm$ 0.07 & 0.32 $\pm$ 0.05 & 0.16 $\pm$ 0.15 & \textbf{0.58 $\pm$ 0.17} \\
\textit{Gesture 4}                                             & \underline{0.34 $\pm$ 0.17} & 0.28 $\pm$ 0.10  & 0.13 $\pm$ 0.08 & 0.22 $\pm$ 0.11 & 0.01 $\pm$ 0.02 & \textbf{0.52 $\pm$ 0.20} \\
\textit{Gesture 5}                                             & \underline{0.40 $\pm$ 0.18} & 0.28 $\pm$ 0.19  & 0.15 $\pm$ 0.03 & 0.21 $\pm$ 0.12 & 0.30 $\pm$ 0.17 &  \textbf{0.53 $\pm$ 0.24} \\
\textit{Gesture 6}                                             & 0.29 $\pm$ 0.14 & 0.28 $\pm$ 0.14  & 0.10 $\pm$ 0.13 & 0.21 $\pm$ 0.13 & \underline{0.35 $\pm$ 0.15} & \textbf{0.43 $\pm$ 0.14}
\end{tabular}
\caption{Gesture class-wise average macro F1 score for a motor-impaired individual with Spinal cord injury in a few-shot continual learning setting (mean$\pm$std). We report the scores after six gestures are trained with five training examples. The best macro F1 score is highlighted in bold whereas the second-best score is underlined.}
\label{tbl1}
\end{table*}

\section{Experiments}
\subsection{Datasets}
\textbf{The SmartWatch Gesture Dataset}~\cite{porzi2013smart,costante2014personalizing} was built for interacting with mobile applications using arm gestures. This dataset contains 20 distinct gestures from eight different subjects. A first-generation Sony smartwatch with a built-in 3-axis accelerometer was worn on the user's right wrist while performing 20 repetitions for each gesture. In total, 3200 sequences were collected and each sequence contains 3-axis acceleration data. We use this dataset as our source domain to build the pre-trained model.
\\
\textbf{The Motion Gesture Dataset}~\cite{vatavu2022understanding} was built to understand the gesture articulation of people with upper-body motor impairments. Six different motion gestures were collected by a group of 12 people (six male and six female) with upper-body motor impairments, ranging ages from 27 to 65 years. The participants had a wide range of disabilities including Spinal cord injury, Traumatic brain injury, Multiple sclerosis, Parkinson's disease, etc. A Samsung Gear Fit 2 smartwatch was used by the participants to collect the wrist gesture's accelerometer data. Each participant repeated each gesture eight times. We utilize this dataset in our few-shot continual learning setting. 

\subsection{Implementation Details}
The sequence length of the data samples varies extensively. As a result, we apply a linear interpolation technique to our source and target datasets so that the sequence length ($L=50$) of each data sample is constant throughout the experiments. While working with the sensor data, outlier features can negatively influence the results. Therefore, dataset standardization is conducted by removing the mean and scaling it according to the interquartile range for each feature. We select 16 out of 20 gestures from the source domain to build a pre-trained model because we do not want any overlapped gestures between the source domain and the target domain. We follow the leave-one-subject-out strategy to pre-train the model. Throughout all experiments, we used the same subjects for a fair comparison. For our architecture, the feature extractor contains one LSTM layer with $64$-dimensional hidden representation, one fully connected layer with $14$ units,  and one dropout layer with a value of $0.5$ between them. A fully connected layer is used as the classification layer. The network architecture remains the same throughout all experiments. The Adam optimizer with learning rate {$10^{-3}$} is used for the few-shot setup. In the few-shot continual learning setting, since each gesture class contains very few training examples, the epoch is set to 15 and the mini-batch contains all examples. We run each experiment 10 times with five different orders of the gesture classes. We report the average accuracy with one, three, and five training examples over all the encountered gesture classes. As this is a continual learning setup, we also report the class-wise macro F1 score and \textit{forgetting} metric to understand each gesture's performance individually. Our memory buffer stores $60\times$ fewer examples compared to traditional replay buffers~\cite{rebuffi2017icarl}.

\begin{figure*}[ht!]
     \centering
     \begin{minipage}[b]{0.39\textwidth}
         \centering
         \includegraphics[width=\textwidth]{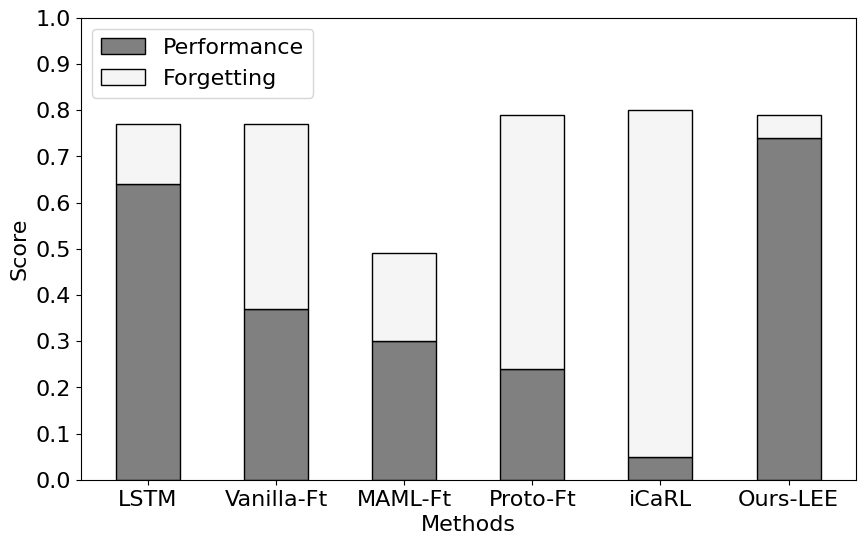}
     \end{minipage}
     \begin{minipage}[b]{0.39\textwidth}
         \centering
         \includegraphics[width=\textwidth]{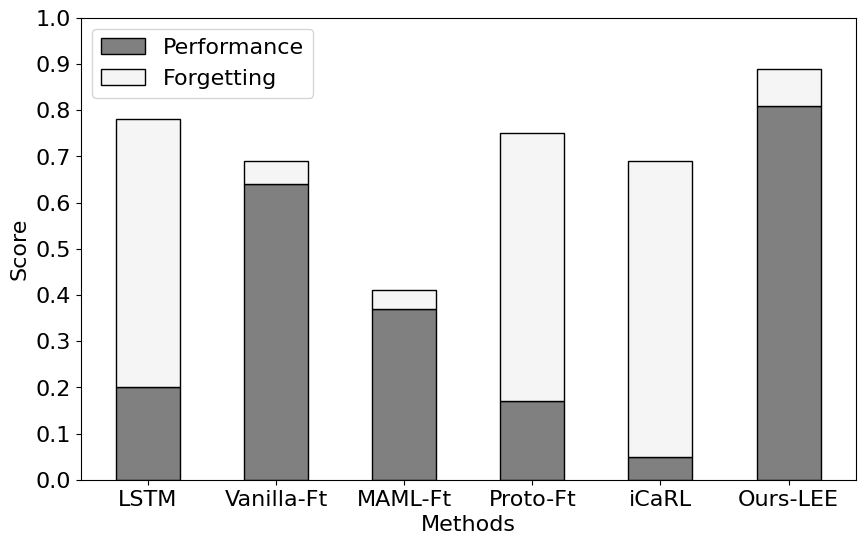}
     \end{minipage}
    \caption{ Performance-\textit{forgetting} scaled score for \textit{Gesture 1} (left) and \textit{Gesture 3} (right) for a motor-impaired individual with Spinal cord injury after six gestures are trained with five training examples.}
    \label{fig4}
    \vspace{-4mm}
\end{figure*}

\subsection{Baselines and Compared Methods}
We compare our methods with different closely related approaches in the few-shot continual learning setting. Since our method utilizes LSTM in the network architecture, we consider comparing it with an LSTM classifier that learns the gesture classes incrementally with a few training examples. Vanilla-Ft, MAML-Ft~\cite{finn2017model}, and Prototypical Net-Ft~\cite{snell2017prototypical} involve fine-tuning the pre-trained models on the few-shot classes. We also compare our method with iCaRL~\cite{rebuffi2017icarl}. We assume that the memory buffer exists in all methods for a fair comparison.

\subsection{Experimental Results}
We compare the average accuracy of the proposed method with other approaches. Figure~\ref{fig3} shows the test accuracies for three participants including a motor-impaired individual with Spinal cord injury, an individual with Parkinson's disease, and an individual with Multiple sclerosis. In most cases, our LEE method outperforms other techniques. We observe that the iCaRL classifier occasionally shows better accuracy than our method while learning two initial gestures. But for the rest of the incrementally added classes, LEE always performs better than the iCaRL classifier. The accuracy of the iCaRL classifier significantly drops for new gesture classes because it fails to capture the diverse and highly variable pattern of unseen gestures with few training examples. The performance increases with the sample size for all methods. Surprisingly, the fine-tuning approach fails compared to the basic LSTM classifier and our LEE.

In a continual learning setup, it is important to perform well in old classes while trained on a new class. Therefore, in Table~\ref{tbl1}, we report the class-wise F1 scores after six gestures are trained with five training examples. Our method always provides a higher macro F1 score than other methods. LEE has a 12.3\% higher F1 score for all gesture classes. Figure~\ref{fig4} shows performance and \textit{forgetting} score for \textit{Gesture 1} and \textit{Gesture 3} after six gestures are trained with five samples. Our LEE method has better performance with less forgetting.

\begin{figure*}[ht!]
     \centering
     \begin{subfigure}[b]{0.30\textwidth}
         \centering
         \includegraphics[width=\textwidth]{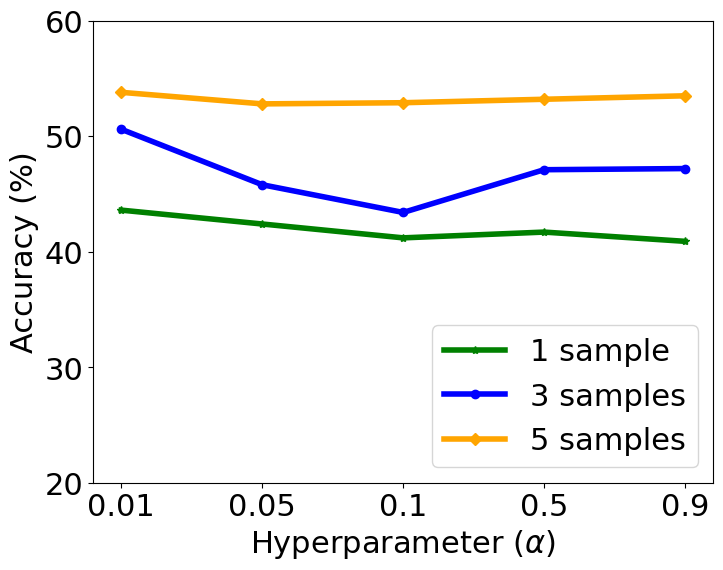}
     \end{subfigure}
     \begin{subfigure}[b]{0.30\textwidth}
         \centering
         \includegraphics[width=\textwidth]{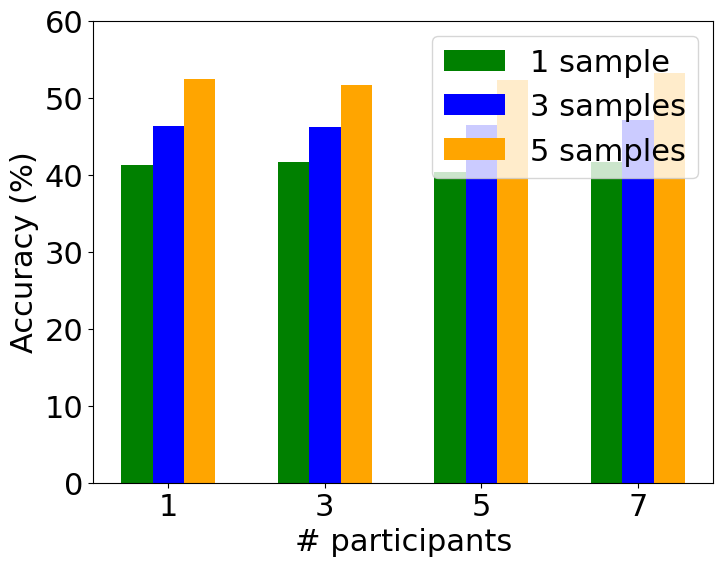}
     \end{subfigure}
     \begin{subfigure}[b]{0.30\textwidth}
         \centering
         \includegraphics[width=\textwidth]{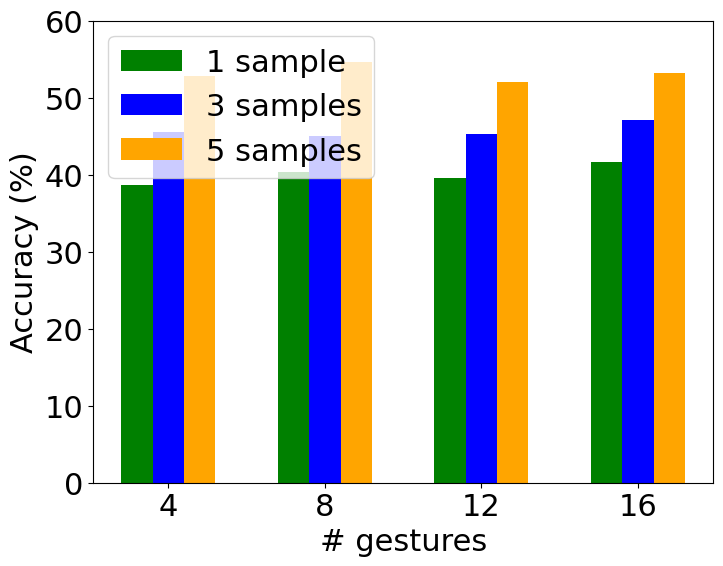}
     \end{subfigure}
    
    \caption{Accuracy for different hyperparameter values (left), number of participants (middle), and number of gestures (right) from the source domain when the preserved latent embedding is produced. We report the accuracy after six gestures are trained with five training examples.}
    \label{fig5}
\end{figure*}

\begin{figure*}[ht!]
     \centering
     \begin{subfigure}[b]{0.30\textwidth}
         \centering
         \includegraphics[width=\textwidth]{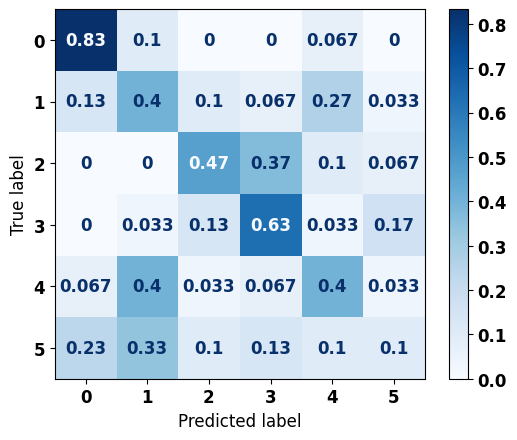}
     \end{subfigure}
     \begin{subfigure}[b]{0.30\textwidth}
         \centering
         \includegraphics[width=\textwidth]{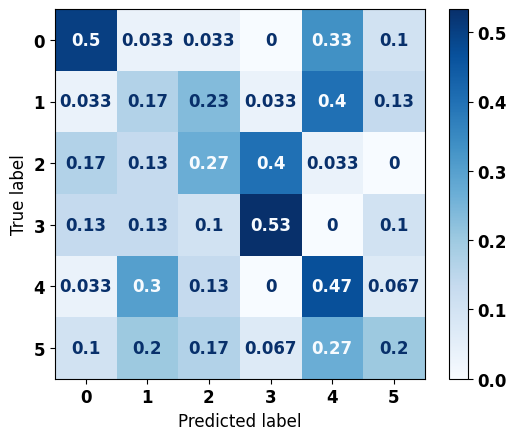}
     \end{subfigure}
     \begin{subfigure}[b]{0.30\textwidth}
         \centering
         \includegraphics[width=\textwidth]{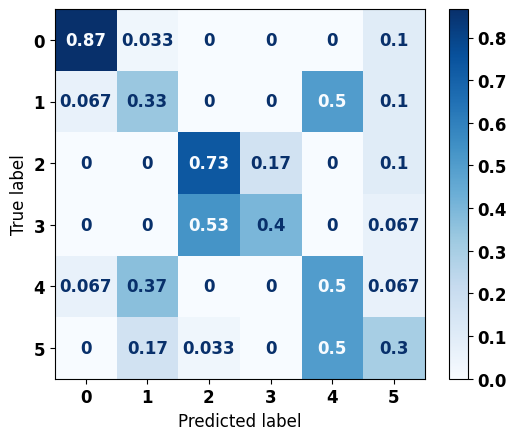}
     \end{subfigure}
    \caption{The confusion matrices for six gesture classes after training with five training examples. left: w/o preserved latent embedding ($\alpha$ = 0); middle: w/o temporary latent embedding ($\alpha$ = 1); right: LEE ($\alpha$ = 0.5).}
    \label{fig6}
    \vspace{-4mm}
\end{figure*}

\section{Ablation Study}
\subsection{Loss Hyperparameter Sensitivity Analysis}
We report the effect of loss hyperparameter sensitivity. We focus only on $\alpha$ as it complements the other hyperparameter ($\beta$) in our continual learning module. We choose $\alpha$ to exploit the \textit{gesture prior knowledge}, selected from $\alpha\in$\{0.01, 0.05, 0.5, 0.1, 0.9\}. According to Figure~\ref{fig5} (left), LEE provides robust accuracy with a wide range of hyperparameters after learning six gestures using five training examples.

\subsection{Number of Participants and Gestures in Source Domain}
We experiment to produce the preserved latent embedding with a different number of participants from the source domain. We achieve higher accuracy with one and three samples using seven different participants (Figure~\ref{fig5} middle). Apart from this, the proposed method is invariant to the number of participants from the source domain. However, it is preferable to utilize a large number of control participants in the source domain to capture a more diversified representation space. We also conduct experiments with a different number of gestures from the source domain to generate the preserved latent embedding (Figure~\ref{fig5} right). Though we get the highest accuracy using 16 gestures, we observe that the accuracy does not change for other different numbers of gestures. Therefore, the observation illustrates that our method is robust to the number of gesture classes.

\subsection{Significance of Embeddings}
We conduct experiments to investigate how the preserved latent embedding and the temporary latent embedding contribute to our proposed method. We apply LEE without one of those embeddings, one at a time, and evaluate the performance. Figure~\ref{fig6} (left) and (middle) show that removing either component of interest results in a less tridiagonal-shaped confusion matrix, indicating a drop in model performance and robustness. We further confirm from Figure~\ref{fig6} (right) that both embeddings jointly contribute to robust performance, and thus these embeddings are the foundations for learning unseen gestures incrementally with limited training examples.

\section{Use Cases and Social Impact}
Wearable sensor-based gesture recognition is becoming popular in many areas including communication, controlling home appliances, and interactive entertainment. As most research doesn't include the motor-impaired population, those individuals face challenges using wearable devices for gesture recognition and communication. We believe our contribution can be impactful for those who need more than a set of pre-defined gestures. Gesture recognition exists for standard movements such as sign language for speech-impaired people but sign language can be difficult to perform for individuals lacking fine motor skills. Our work is part of a fast and flexible gesture-to-speech recognition system that we are developing in collaboration with Shirley Ryan AbilityLab~\footnote{https://www.sralab.org/}. The need for such solutions is underscored by the prevalence of motor impairments that also impact speech. $12.1\%$ of the population has a motor disability~\cite{CDC_Disability} while $7.6\%$ have a speech disorder~\cite{NIDCD_Stats} through Stroke ($795,000$ cases each year), Parkinson's disease ($1$ million), Multiple sclerosis ($727,000$), Spinal cord injury ($294,000$) and Cerebral palsy ($764,000$)~\cite{wallin2019prevalence,white2016spinal}. The proposed method has the potential to transform the lives of these individuals by providing a more natural and efficient mode of communication, improving quality of life, enhancing interaction, and reducing the burden on caregivers.    

\vspace{-2mm}
\section{Conclusion and Future Work}
Hand gestures are natural and flexible means of communication. Available wearable sensor-based hand gesture solutions are widely adopted for the normal population and these solutions fail to capture highly variable and inconsistent data samples from motor-impaired people. Moreover, in the real world, motor-impaired individuals face challenges in performing predefined gestures, and many valuable use cases rely on acquiring new gestures. However, 
a substantial amount of data samples is needed to develop a strong hand gesture recognition method. Therefore, we introduce a novel method called Latent Embedding Exploitation (LEE) to learn novel gesture classes incrementally using a few samples from motor-impaired individuals. We experimentally show that our method outperforms the existing baselines. Our method helps motor-impaired persons leverage wearable devices and their unique movement styles can be learned and applied in human-computer interaction and social communication. By enabling meaningful interactions with motor-impaired individuals and seamlessly integrating wearable devices into their daily lives, we open the gate to collecting invaluable data from this underrepresented group in real-world scenarios. This data collection paradigm can play a central role in facilitating the advancements in other machine learning research, benefiting not only motor-impaired individuals but also contributing to broader technological innovation. In the future, we will integrate our method with a wearable application and online learning will be explored to assist motor-impaired individuals to input their custom, flexible gestures in real-time. Furthermore, we will survey to collect the opinions of the population and tailor our approach accordingly.




\section*{Acknowledgments}
Research reported in this publication was supported by the Eunice Kennedy Shriver National Institute of Child Health \& Human Development of the National Institutes of Health under Award Number P2CHD101899.

\bibliographystyle{named}
\bibliography{ijcai24}

\end{document}


\onecolumn
\setcounter{section}{7}                   
\renewcommand\thetable{\arabic{table}}
\section{Appendix} 
\subsection{Additional Results}
We report the confusion matrices for five different orders after six gestures are trained with five training examples. The hand gestures are known as `tap', `double tap', `circle', `rotate fast and slow', `rotate slow and fast', and `shake'.   Figure~\ref{fig7} - ~\ref{fig11} illustrates that the LEE method has a more robust performance than other methods. For certain orders, LEE mistakes one gesture for another. Since order is an influential factor, we intend to explore it further in our future work. 

\begin{figure}[ht!]
    \centering
     \begin{minipage}[b]{0.25\textwidth}
        \centering
        \includegraphics[width=\textwidth]{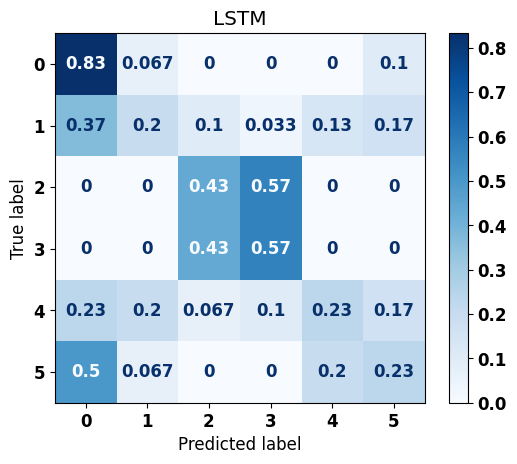}
     \end{minipage}
     \hspace{1mm}
     \begin{minipage}[b]{0.25\textwidth}
         \centering
         \includegraphics[width=\textwidth]{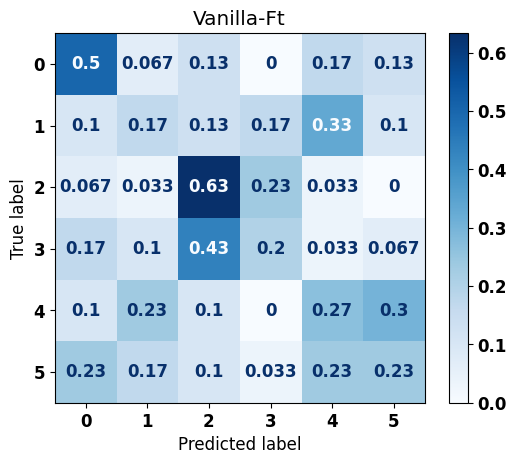}
     \end{minipage}
     \hspace{1mm}
     \begin{minipage}[b]{0.25\textwidth}
         \centering
         \includegraphics[width=\textwidth]{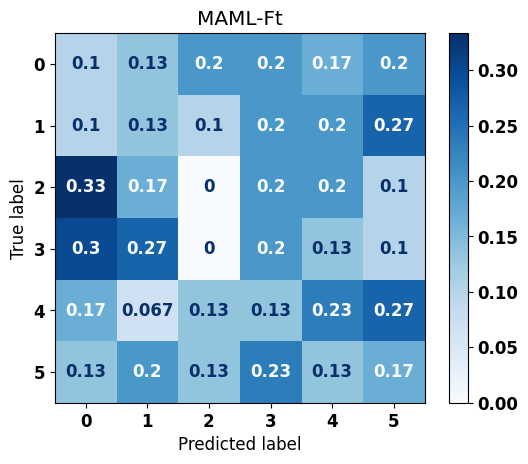}
     \end{minipage}
     \begin{minipage}[b]{0.25\textwidth}
         \centering
         \includegraphics[width=\textwidth]{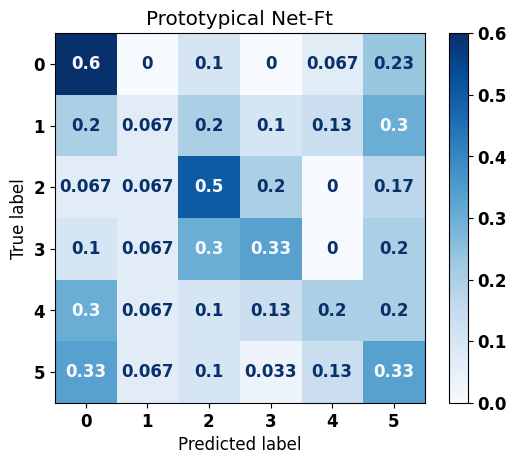}
     \end{minipage}
     \hspace{1mm}
     \begin{minipage}[b]{0.25\textwidth}
         \centering
         \includegraphics[width=\textwidth]{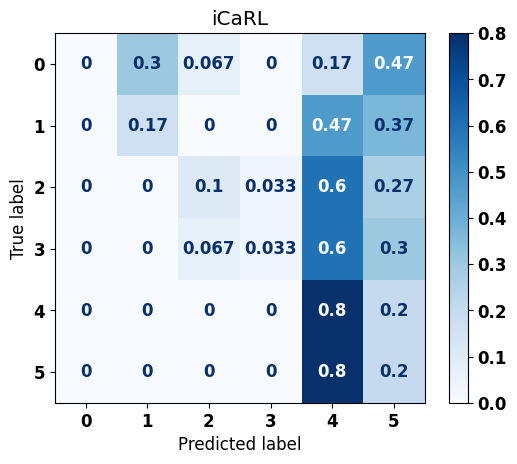}
     \end{minipage}
     \hspace{1mm}
     \begin{minipage}[b]{0.25\textwidth}
         \centering
         \includegraphics[width=\textwidth]{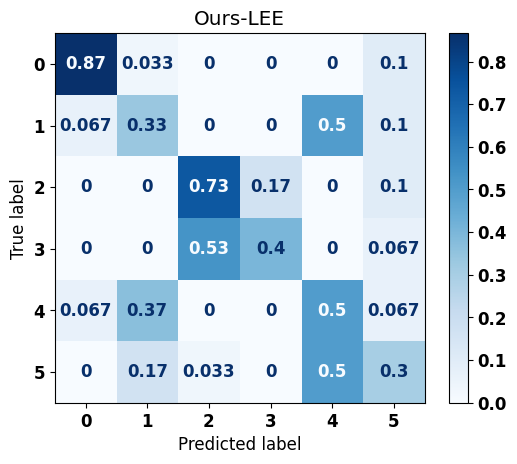}
     \end{minipage}
    \setcounter{figure}{6}                   
    \renewcommand\thetable{\arabic{table}} 
    \caption{Order 1: `circle', `double tap', `rotate fast and slow', `rotate slow and fast', `shake', `tap'. The confusion matrices for six gesture classes after training with five training examples.}
    \label{fig7}
\end{figure}

\begin{figure}[ht!]
     \centering
     \begin{minipage}[b]{0.25\textwidth}
         \centering
         \includegraphics[width=\textwidth]{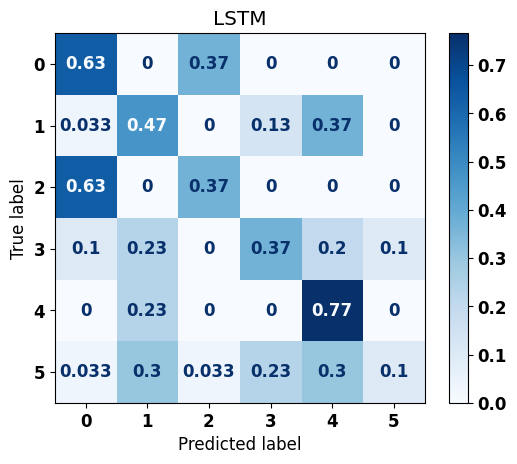}
     \end{minipage}
     \hspace{1mm}
     \begin{minipage}[b]{0.25\textwidth}
         \centering
         \includegraphics[width=\textwidth]{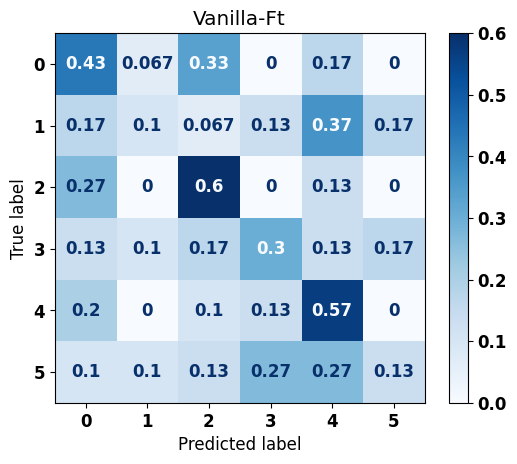}
     \end{minipage}
     \hspace{1mm}
     \begin{minipage}[b]{0.25\textwidth}
         \centering
         \includegraphics[width=\textwidth]{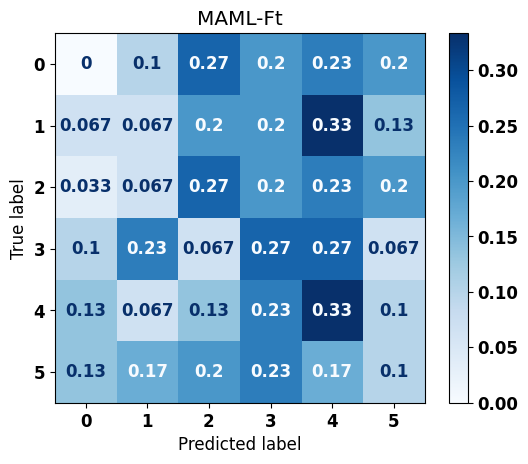}
     \end{minipage}
     \hspace{1mm}
     \begin{minipage}[b]{0.25\textwidth}
         \centering
         \includegraphics[width=\textwidth]{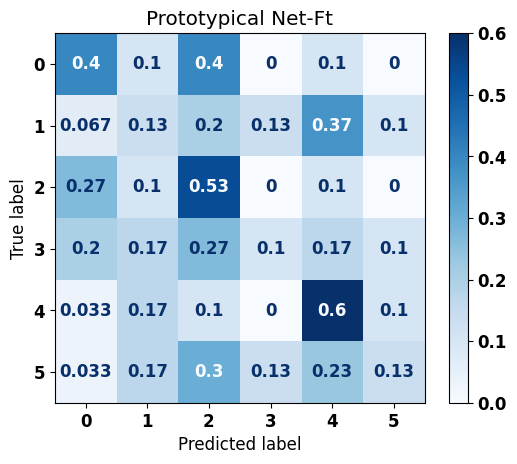}
     \end{minipage}
     \hspace{1mm}
     \begin{minipage}[b]{0.25\textwidth}
         \centering
         \includegraphics[width=\textwidth]{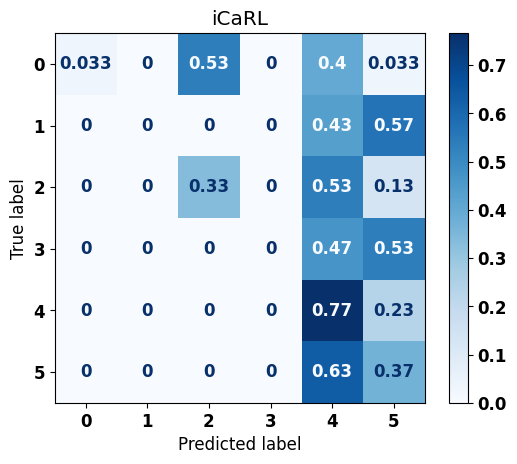}
     \end{minipage}
     \hspace{1mm}
     \begin{minipage}[b]{0.25\textwidth}
         \centering
         \includegraphics[width=\textwidth]{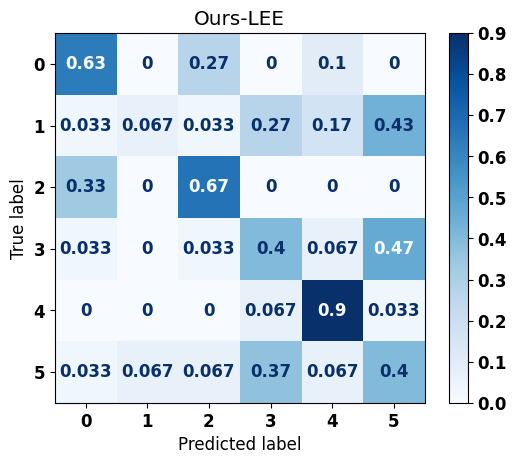}
     \end{minipage}
    \setcounter{figure}{7}                   
    \renewcommand\thetable{\arabic{table}} 
    \caption{Order 2: `rotate slow and fast', `tap', `rotate fast and slow', `shake', `circle', `double tap'. The confusion matrices for six gesture classes after training with five training examples.}
    \label{fig8}
\end{figure}

\begin{figure}[ht!]
     \centering
     \begin{minipage}[b]{0.25\textwidth}
         \centering
         \includegraphics[width=\textwidth]{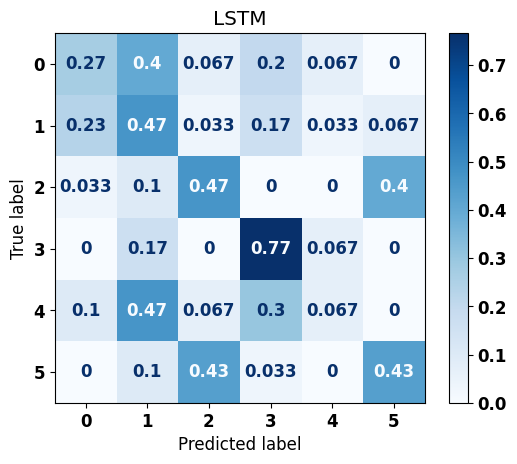}
     \end{minipage}
     \hspace{1mm}
     \begin{minipage}[b]{0.25\textwidth}
         \centering
         \includegraphics[width=\textwidth]{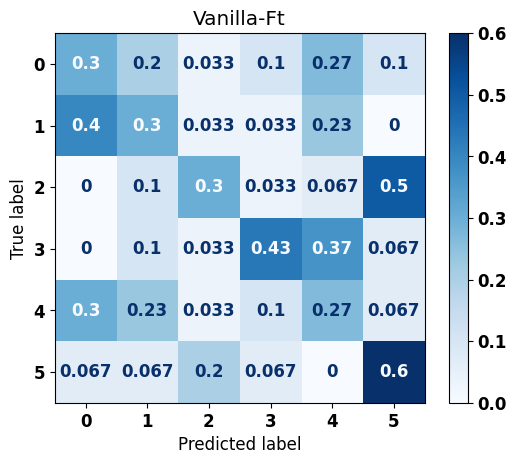}
     \end{minipage}
     \hspace{1mm}
     \begin{minipage}[b]{0.25\textwidth}
         \centering
         \includegraphics[width=\textwidth]{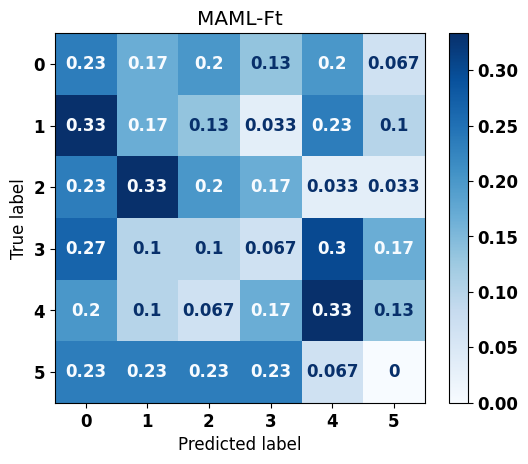}
     \end{minipage}
     \hspace{1mm}
     \begin{minipage}[b]{0.25\textwidth}
         \centering
         \includegraphics[width=\textwidth]{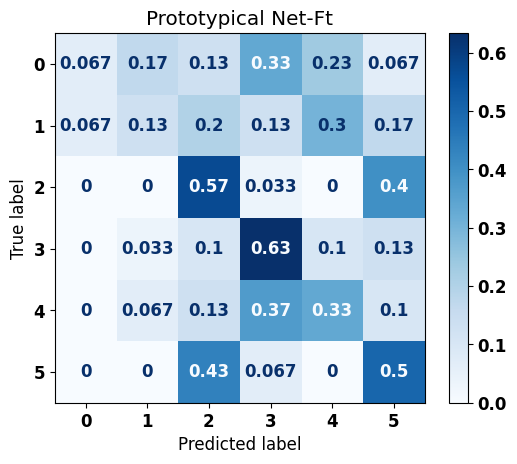}
     \end{minipage}
     \hspace{1mm}
     \begin{minipage}[b]{0.25\textwidth}
         \centering
         \includegraphics[width=\textwidth]{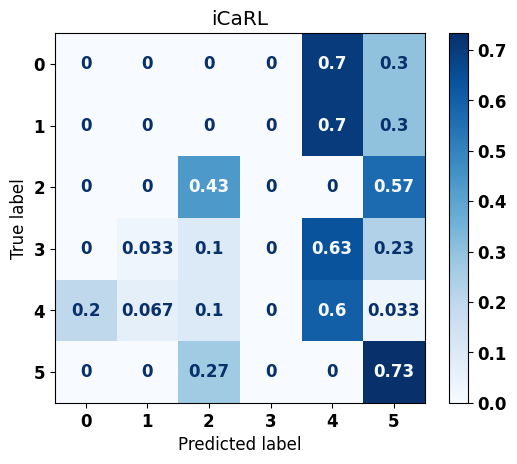}
     \end{minipage}
     \hspace{1mm}
     \begin{minipage}[b]{0.25\textwidth}
         \centering
         \includegraphics[width=\textwidth]{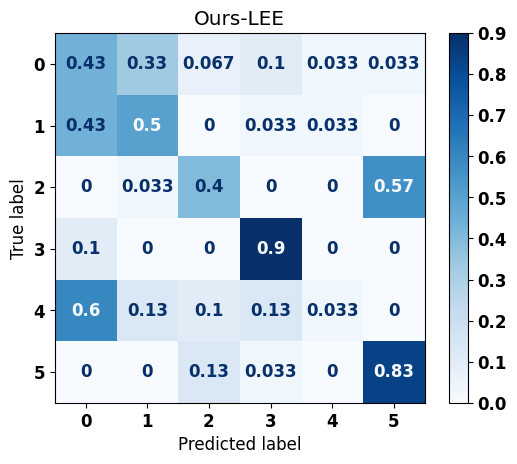}
     \end{minipage}
    \setcounter{figure}{8}                   
    \renewcommand\thetable{\arabic{table}} 
    \caption{Order 3: `double tap', `shake', `rotate fast and slow', `circle', `tap', `rotate slow and fast'. The confusion matrices for six gesture classes after training with five training examples.}
    \label{fig9}
\end{figure}

\newpage

\begin{figure}[ht!]
     \centering
     \begin{minipage}[b]{0.25\textwidth}
         \centering
         \includegraphics[width=\textwidth]{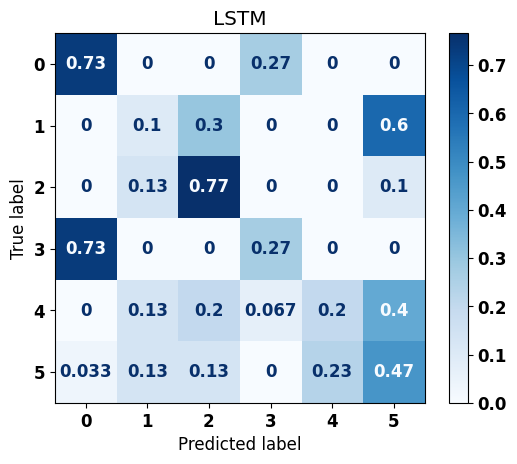}
     \end{minipage}
     \hspace{1mm}
     \begin{minipage}[b]{0.25\textwidth}
         \centering
         \includegraphics[width=\textwidth]{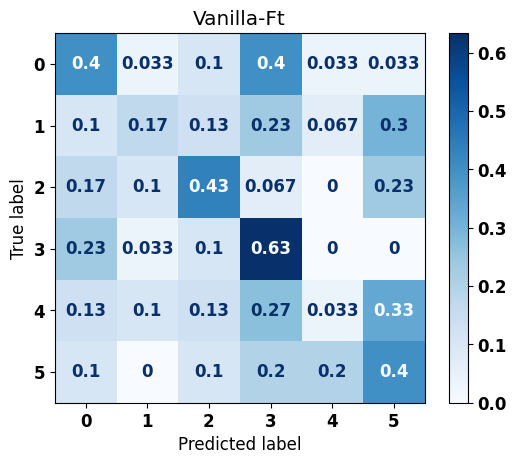}
     \end{minipage}
     \hspace{1mm}
     \begin{minipage}[b]{0.25\textwidth}
         \centering
         \includegraphics[width=\textwidth]{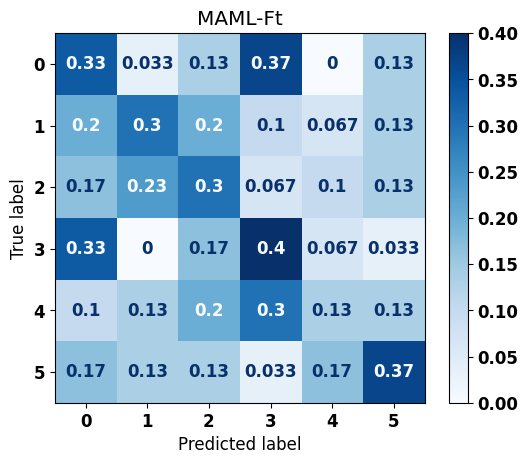}
     \end{minipage}
     \hspace{1mm}
     \begin{minipage}[b]{0.25\textwidth}
         \centering
         \includegraphics[width=\textwidth]{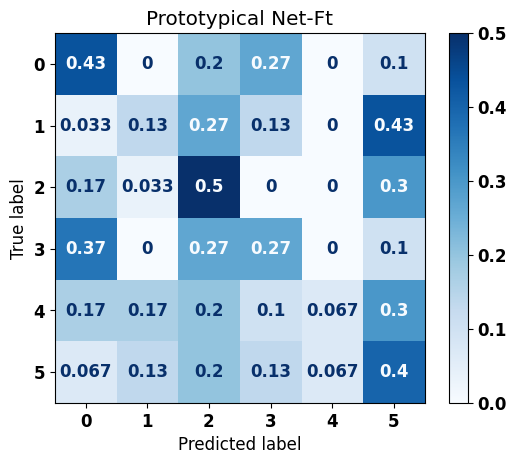}
     \end{minipage}
     \hspace{1mm}
     \begin{minipage}[b]{0.25\textwidth}
         \centering
         \includegraphics[width=\textwidth]{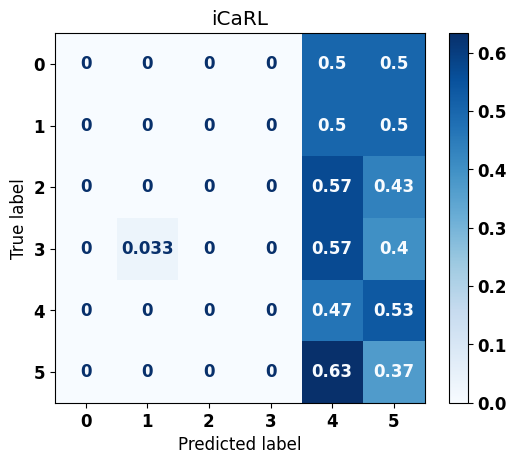}
     \end{minipage}
     \hspace{1mm}
     \begin{minipage}[b]{0.25\textwidth}
         \centering
         \includegraphics[width=\textwidth]{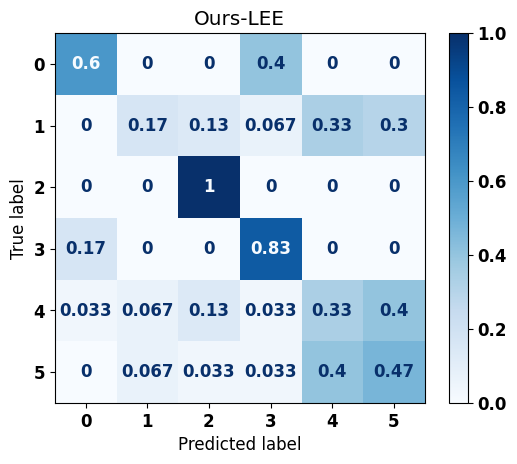}
     \end{minipage}
    \setcounter{figure}{9}                   
    \renewcommand\thetable{\arabic{table}} 
    \caption{Order 4: `rotate fast and slow', `tap', `circle', `rotate slow and fast', `double tap', `shake'.}
    \label{fig10}
    \vspace{40mm}
\end{figure}

\begin{figure}[ht!]
     \centering
     \begin{minipage}[b]{0.25\textwidth}
         \centering
         \includegraphics[width=\textwidth]{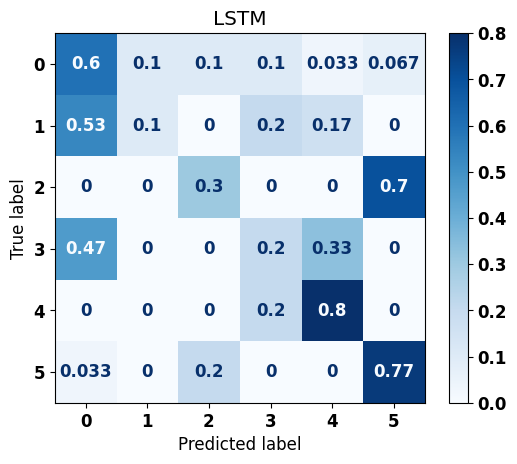}
     \end{minipage}
     \hspace{1mm}
     \begin{minipage}[b]{0.25\textwidth}
         \centering
         \includegraphics[width=\textwidth]{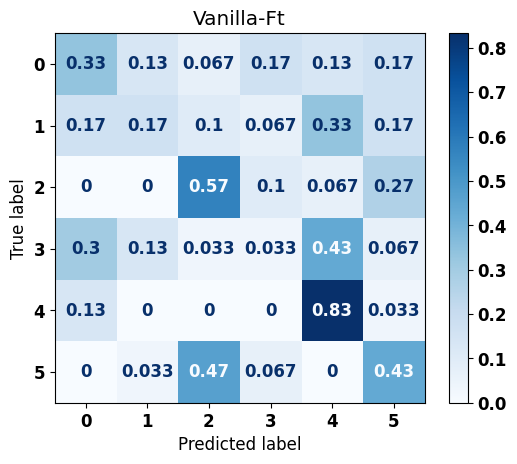}
     \end{minipage}
     \hspace{1mm}
     \begin{minipage}[b]{0.25\textwidth}
         \centering
         \includegraphics[width=\textwidth]{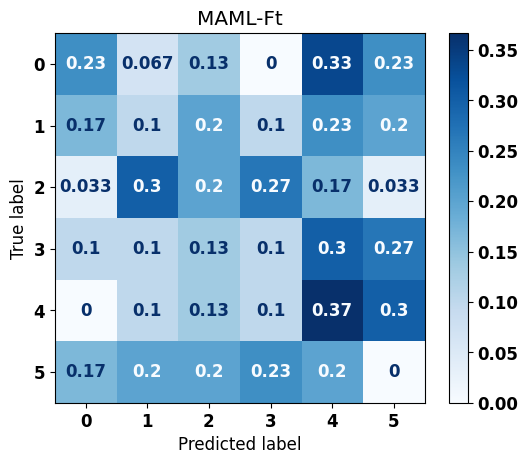}
     \end{minipage}
     \hspace{1mm}
     \begin{minipage}[b]{0.25\textwidth}
         \centering
         \includegraphics[width=\textwidth]{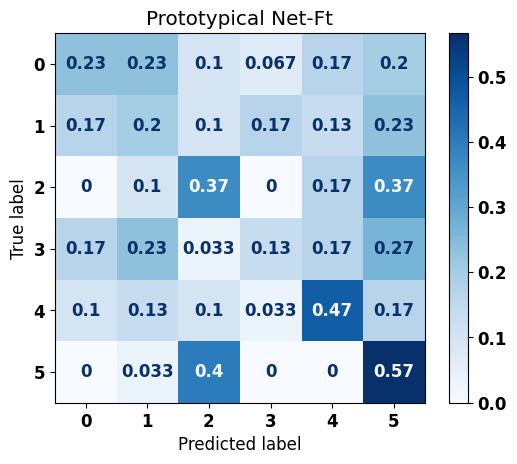}
     \end{minipage}
     \hspace{1mm}
     \begin{minipage}[b]{0.25\textwidth}
         \centering
         \includegraphics[width=\textwidth]{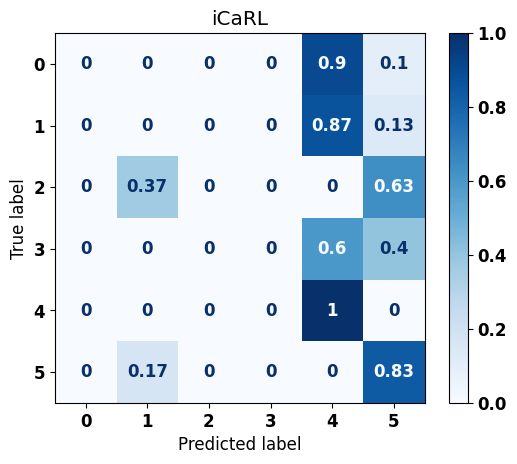}
     \end{minipage}
     \hspace{1mm}
     \begin{minipage}[b]{0.25\textwidth}
         \centering
         \includegraphics[width=\textwidth]{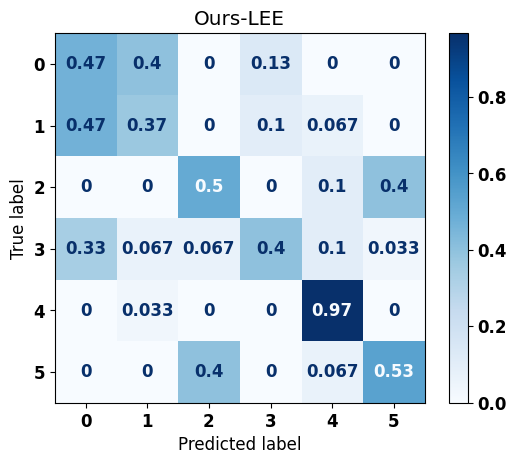}
     \end{minipage}
    \setcounter{figure}{10}                   
    \renewcommand\thetable{\arabic{table}} 
    \caption{Order 5: `shake', `double tap', `rotate slow and fast', `tap', `circle', `rotate fast and slow'.}
    \label{fig11}
\end{figure}

We report overall average accuracy for motor-impaired individuals with Spinal cord injury, Parkinson's disease, and Multiple sclerosis after training with one, three, and five training examples (Table~\ref{tbl2}-\ref{tbl4}). We also report Gesture class-wise average macro F1-score for individuals with Parkinson's disease (Table~\ref{tbl5}) and Multiple sclerosis (Table~\ref{tbl6}). As our method is implemented in a few-shot continual learning setting, Table~\ref{tbl7} shows the \textit{forgetting} score for each gesture after training with five training examples. Our method always outperforms other approaches.


\begin{table}[ht!]
\centering
\begin{tabular}{c|c|cccccc}
\hline
\multirow{2}{*}{\# gestures} & \multirow{2}{*}{\# samples} & \multicolumn{6}{c}{\textbf{Methods}}                                                                    \\ \cline{3-8} 
                             &                             & LSTM            & Vanilla-Ft & MAML-Ft  & Prototypical Net-Ft & iCaRL              & Ours-LEE           \\ \hline
\multirow{3}{*}{2}           & 1                           & \underline{72.0 $\pm$ 18.0} & 61.0 $\pm$ 6.4   & 48.0 $\pm$ 6.3 & 67.0 $\pm$ 8.8            & \textbf{77.3 $\pm$ 23.2} & 66.0 $\pm$ 8.4           \\
                             & 3                           & 77.4 $\pm$ 19.2 & 71.0 $\pm$ 14.9  & 58.0 $\pm$ 7.0 & 76.3 $\pm$ 10.9           & \textbf{83.7 $\pm$ 19.3} & \underline{77.6 $\pm$ 18.9}          \\
                             & 5                           & 76.1 $\pm$ 18.2       & 78.3 $\pm$ 15.9  & 56.3 $\pm$ 4.0 & 78.1 $\pm$ 9.1            & \textbf{82.4 $\pm$ 18.2} & \underline{80.7 $\pm$ 18.9}    \\ \hline
\multirow{3}{*}{3}           & 1                           & \underline{65.1 $\pm$ 3.6}  & 54.1 $\pm$ 7.6   & 38.1 $\pm$ 7.8 & 58.5 $\pm$ 9.5            & -                  & \textbf{68.9 $\pm$ 7.3} \\
                             & 3                           & \underline{72.9 $\pm$ 4.9}  & 63.8 $\pm$ 4.1   & 39.9 $\pm$ 4.6 & 61.9 $\pm$ 5.7            & -                  & \textbf{74.7 $\pm$ 9.5} \\
                             & 5                           & \underline{70.7 $\pm$ 5.4}  & 66.8 $\pm$ 11.1  & 44.6 $\pm$ 4.1 & 66.0 $\pm$ 8.0            & -                  & \textbf{81.1 $\pm$ 10.3} \\ \hline
\multirow{3}{*}{4}           & 1                           & \underline{51.0\ $\pm$ 10.2} & 38.5 $\pm$ 7.4   & 25.0 $\pm$ 1.4 & 44.7 $\pm$ 4.7            & 49.5 $\pm$ 6.3           & \textbf{55.9 $\pm$ 6.7}  \\
                             & 3                           & \underline{52.3 $\pm$ 2.9} & 47.7 $\pm$ 4.8   & 27.8 $\pm$ 3.0 & 47.9 $\pm$ 2.1            & 44.5 $\pm$ 6.0           & \textbf{64.7 $\pm$ 6.5}  \\
                             & 5                           & \underline{55.8 $\pm$ 5.3} & 54.1 $\pm$ 7.6   & 28.3 $\pm$ 2.6 & 53.3 $\pm$ 7.6            & 45.8 $\pm$ 8.6           & \textbf{67.1 $\pm$ 8.4}  \\ \hline
\multirow{3}{*}{5}           & 1                           & \underline{45.1 $\pm$ 3.1} & 38.7 $\pm$ 5.0   & 21.9 $\pm$ 2.4 & 32.3 $\pm$ 4.4            & -                  & \textbf{52.4 $\pm$ 2.4}  \\
                             & 3                           & \underline{47.7 $\pm$ 2.6} & 36.8 $\pm$ 3.8   & 21.6 $\pm$ 5.2 & 34.0 $\pm$ 3.0            & -                  & \textbf{59.1 $\pm$ 4.4}  \\
                             & 5                           & \underline{51.0 $\pm$ 0.9} & 43.7 $\pm$ 4.2   & 23.5 $\pm$ 4.7 & 37.0 $\pm$ 6.5            & -                  & \textbf{64.2 $\pm$ 4.9}  \\ \hline
\multirow{3}{*}{6}           & 1                           & 35.8 $\pm$ 3.2        & 27.7 $\pm$ 2.7   & 20.7 $\pm$ 3.0 & 26.2 $\pm$ 3.1            & \underline{39.3 $\pm$ 6.5}     & \textbf{41.7 $\pm$ 1.6}  \\
                             & 3                           & \underline{38.5 $\pm$ 2.8}  & 33.5 $\pm$ 3.8   & 21.1 $\pm$ 4.1 & 29.6 $\pm$ 3.7            & 35.3 $\pm$ 3.8           & \textbf{47.1 $\pm$ 3.4}  \\
                             & 5                           & \underline{43.1 $\pm$ 1.9}  & 35.8 $\pm$ 2.1   & 19.1 $\pm$ 5.9 & 33.1 $\pm$ 2.5            & 24.1 $\pm$ 6.0           & \textbf{53.2 $\pm$ 2.0} 
\end{tabular}
\setcounter{table}{1}                   
\renewcommand\thetable{\arabic{table}} 
\caption{Test accuracies for a motor-impaired individual with Spinal cord injury in a few-shot continual learning setting (mean$\pm$std). The accuracy represents the total accuracy over all the gesture classes encountered trained with one, three, and five samples. The best accuracy is highlighted in bold whereas the second-best accuracy is underlined.}
\label{tbl2}
\vspace{30mm}
\end{table}

\begin{table}[ht!]
\centering
\begin{tabular}{c|c|cccccc}
\hline
\multirow{2}{*}{\# gestures} & \multirow{2}{*}{\# samples} & \multicolumn{6}{c}{\textbf{Methods}}                                                                        \\ \cline{3-8} 
                             &                             & LSTM            & Vanilla-Ft     & MAML-Ft  & Prototypical Net-Ft & iCaRL              & Ours-LEE           \\ \hline
\multirow{3}{*}{2}           & 1                           & 55.0 $\pm$ 18.6       & 53.6 $\pm$ 13.5      & 46.0 $\pm$ 4.9 & 57.6 $\pm$ 9.9            & \textbf{64.6 $\pm$ 19.7}    & \underline{62.0 $\pm$ 12.1}  \\
                             & 3                           & 59.7 $\pm$ 19.9       & 59.3 $\pm$ 7.9       & 48.7 $\pm$ 8.3 & 60.3 $\pm$ 13.1           & \textbf{81.6 $\pm$ 10.0} & \underline{64.6 $\pm$ 10.3}    \\
                             & 5                           & \underline{68.7 $\pm$ 17.6} & 61.6 $\pm$ 13.5      & 57.1 $\pm$ 4.2 & 67.4 $\pm$ 7.4            & \textbf{85.6 $\pm$ 10.0} & 64.3 $\pm$ 14.2          \\ \hline
\multirow{3}{*}{3}           & 1                           & 45.5 $\pm$ 5.4        & 45.1 $\pm$ 4.2       & 37.4 $\pm$ 5.0 & \underline{47.5 $\pm$ 9.3}      & -                  & \textbf{55.4 $\pm$ 4.7}  \\
                             & 3                           & \underline{51.9 $\pm$ 13.8} & 48.2 $\pm$ 3.7       & 38.1 $\pm$ 7.7 & 50.2 $\pm$ 4.7            & -                  & \textbf{59.7 $\pm$ 8.9} \\
                             & 5                           & \underline{57.8 $\pm$ 12.5} & 52.0 $\pm$ 5.3       & 37.0 $\pm$ 6.3 & 51.1 $\pm$ 7.2            & -                  & \textbf{63.5 $\pm$ 7.6}  \\ \hline
\multirow{3}{*}{4}           & 1                           & 33.7 $\pm$ 5.5        & 35.1 $\pm$ 3.5       & 24.9 $\pm$ 2.7 & 33.9 $\pm$ 7.0            & \underline{44.6 $\pm$ 6.9}     & \textbf{47.0 $\pm$ 5.2}  \\
                             & 3                           & 38.9 $\pm$ 8.8        & 36.3 $\pm$ 5.8       & 27.3 $\pm$ 2.4 & 35.9 $\pm$ 3.4            & \underline{43.4 $\pm$ 9.7}     & \textbf{55.3 $\pm$ 11.3}  \\
                             & 5                           & \underline{46.0 $\pm$ 7.4}  & 37.5 $\pm$ 4.3       & 27.8 $\pm$ 1.9 & 37.8 $\pm$ 5.3            & 43.9 $\pm$ 3.3           & \textbf{52.4 $\pm$ 4.3}  \\ \hline
\multirow{3}{*}{5}           & 1                           & 24.8 $\pm$ 3.7        & \underline{30.1 $\pm$ 2.8} & 20.1 $\pm$ 4.9 & 28.1 $\pm$ 1.9            & -                  & \textbf{41.2 $\pm$ 3.3}  \\
                             & 3                           & \underline{34.1 $\pm$ 4.0}  & 33.7 $\pm$ 2.4       & 22.5 $\pm$ 3.4 & 30.4 $\pm$ 2.7            & -                  & \textbf{45.6 $\pm$ 7.8}  \\
                             & 5                           & \underline{37.7 $\pm$ 1.9}  & 33.2 $\pm$ 5.1       & 22.9 $\pm$ 1.7 & 32.6 $\pm$ 4.2            & -                  & \textbf{47.8 $\pm$ 2.3}  \\ \hline
\multirow{3}{*}{6}           & 1                           & 21.9 $\pm$ 1.3        & \underline{28.7 $\pm$ 2.9} & 16.1 $\pm$ 2.9 & 24.9 $\pm$ 1.9            & 22.3 $\pm$ 5.5           & \textbf{36.3 $\pm$ 3.0}  \\
                             & 3                           & 25.8 $\pm$ 3.1        & \underline{27.0 $\pm$ 2.5} & 19.1 $\pm$ 3.1 & 22.1 $\pm$ 2.7            & 24.1 $\pm$ 9.6           & \textbf{43.5 $\pm$ 3.3}  \\
                             & 5                           & \underline{30.9 $\pm$ 1.7}  & 23.5 $\pm$ 3.3       & 19.8 $\pm$ 2.8 & 24.1 $\pm$ 1.7            & 22.5 $\pm$ 3.7           & \textbf{40.2 $\pm$ 2.1} 
\end{tabular}
\setcounter{table}{2}                   
\renewcommand\thetable{\arabic{table}} 
\caption{Test accuracies for a motor-impaired individual with Parkinson's disease in a few-shot continual learning setting (mean$\pm$std). The accuracy represents the total accuracy over all the gesture classes encountered trained with one, three, and five samples. The best accuracy is highlighted in bold whereas the second-best accuracy is underlined.}
\label{tbl3}
\end{table}


\begin{table}[ht!]
\centering
\begin{tabular}{c|c|cccccc}
\hline
\multirow{2}{*}{\# gestures} & \multirow{2}{*}{\# samples} & \multicolumn{6}{c}{\textbf{Methods}}                                                                         \\ \cline{3-8} 
                             &                             & LSTM           & Vanilla-Ft      & MAML-Ft   & Prototypical Net-Ft & iCaRL              & Ours-LEE           \\ \hline
\multirow{3}{*}{2}           & 1                           & 42.6 $\pm$ 14.6      & 56.3 $\pm$ 9.5        & 50.0 $\pm$ 5.2  & \underline{60.6 $\pm$ 16.0}     & 54.3 $\pm$ 13.8          & \textbf{65.7 $\pm$ 12.9}  \\
                             & 3                           & 72.9 $\pm$ 5.7 & 59.4 $\pm$ 15.3       & 53.3 $\pm$ 8.2  & 71.3 $\pm$ 7.7            & \textbf{81.2 $\pm$ 8.9}  & \underline{74.3 $\pm$ 4.5}          \\
                             & 5                           & 69.5 $\pm$ 12.8      & 62.3 $\pm$ 11.3       & 54.3 $\pm$ 6.0  & \underline{73.3 $\pm$ 4.8}            & \textbf{82.6 $\pm$ 11.1} & 72.9 $\pm$ 5.9     \\ \hline
\multirow{3}{*}{3}           & 1                           & 48.0 $\pm$ 13.0      & \underline{50.9 $\pm$ 10.8} & 41.9 $\pm$ 3.9  & 49.1 $\pm$ 11.3           & -                  & \textbf{56.3 $\pm$ 9.2}  \\
                             & 3                           & \underline{59.2 $\pm$ 9.1} & 55.2 $\pm$ 8.5        & 40.1 $\pm$ 3.2  & 51.5 $\pm$ 12.7           & -                  & \textbf{67.1 $\pm$ 3.9}  \\
                             & 5                           & \underline{60.4 $\pm$ 6.4} & 56.1 $\pm$ 8.4        & 41.1 $\pm$ 11.0 & 54.3 $\pm$ 5.0            & -                  & \textbf{73.1 $\pm$ 8.9}  \\ \hline
\multirow{3}{*}{4}           & 1                           & \underline{44.5 $\pm$ 9.6}       & 39.7 $\pm$ 6.7        & 28.6 $\pm$ 3.0  & 41.3 $\pm$ 6.5            & 42.0 $\pm$ 13.4    & \textbf{48.5 $\pm$ 9.9}  \\
                             & 3                           & \underline{51.4 $\pm$ 6.1} & 45.9 $\pm$ 11.5       & 30.0 $\pm$ 2.1  & 38.7 $\pm$ 8.2            & 41.5 $\pm$ 3.1           & \textbf{60.6 $\pm$ 9.2} \\
                             & 5                           & \underline{44.9 $\pm$ 3.9} & 43.5 $\pm$ 8.3        & 28.4 $\pm$ 3.5  & 43.3 $\pm$ 2.3            & 38.5 $\pm$ 7.7           & \textbf{59.5 $\pm$ 8.6}  \\ \hline
\multirow{3}{*}{5}           & 1                           & \underline{31.5 $\pm$ 6.4} & 28.3 $\pm$ 4.7        & 23.7 $\pm$ 2.0  & 30.0 $\pm$ 3.5            & -                  & \textbf{39.5 $\pm$ 5.8}  \\
                             & 3                           & 35.8 $\pm$ 2.1       & \underline{36.8 $\pm$ 4.0}  & 24.6 $\pm$ 4.9  & 35.0 $\pm$ 3.0            & -                  & \textbf{46.2 $\pm$ 6.9}  \\
                             & 5                           & 38.0 $\pm$ 4.6       & \underline{38.4 $\pm$ 5.8}  & 22.7 $\pm$ 4.4  & 30.5 $\pm$ 2.9            & -                  & \textbf{52.0 $\pm$ 7.0}  \\ \hline
\multirow{3}{*}{6}           & 1                           & \underline{26.5 $\pm$ 2.7} & 26.3 $\pm$ 2.1        & 17.5 $\pm$ 2.5  & 26.1 $\pm$ 2.0            & 21.4 $\pm$ 4.7           & \textbf{32.7 $\pm$ 2.9}  \\
                             & 3                           & \underline{31.2 $\pm$ 3.4} & 31.0 $\pm$ 3.1        & 22.3 $\pm$ 4.4  & 27.4 $\pm$ 4.0            & 29.1 $\pm$ 7.8           & \textbf{40.7 $\pm$ 3.8}  \\
                             & 5                           & 31.5 $\pm$ 4.1       & \underline{33.1 $\pm$ 3.9}  & 20.3 $\pm$ 4.1  & 23.2 $\pm$ 3.1            & 23.2 $\pm$ 7.4           & \textbf{44.3 $\pm$ 3.5} 
\end{tabular}
\setcounter{table}{3}         
\renewcommand\thetable{\arabic{table}}
\caption{Test accuracies for a motor-impaired individual with Multiple sclerosis in a few-shot continual learning setting (mean$\pm$std). The accuracy represents the total accuracy over all the gesture classes encountered trained with one, three, and five samples. The best accuracy is highlighted in bold whereas the second-best accuracy is underlined.}
\label{tbl4}
\end{table}


\begin{table}[ht!]
\centering
\small
\begin{tabular}{c|cccccc}
\hline
\multicolumn{1}{c|}{\multirow{2}{*}{Gesture classes}} & \multicolumn{6}{c}{\textbf{Methods}}                                          \\ \cline{2-7} 
\multicolumn{1}{c|}{}                                 & LSTM      & Vanilla-Ft & MAML-Ft   & Prototypical Net-Ft & iCaRL & \textbf{Ours-LEE} \\ \hline
\textit{Gesture 1}                                             & \underline{0.22 $\pm$ 0.16} & 0.17 $\pm$ 0.08  & 0.18 $\pm$ 0.05 & 0.19 $\pm$ 0.08 & 0.12 $\pm$ 0.10 & \textbf{0.37 $\pm$ 0.11} \\
\textit{Gesture 2}                                             & \underline{0.21 $\pm$ 0.05} & 0.15 $\pm$ 0.04  & 0.17 $\pm$ 0.04 & 0.12 $\pm$ 0.10 & 0.01 $\pm$ 0.01 & \textbf{0.22 $\pm$ 0.13} \\
\textit{Gesture 3}                                             &      
\underline{0.26 $\pm$ 0.15} & 0.23 $\pm$ 0.09  & 0.13 $\pm$ 0.07 & 0.19 $\pm$ 0.02 & 0.02 $\pm$ 0.15 & \textbf{0.41 $\pm$ 0.14} \\
\textit{Gesture 4}                                             & \underline{0.32 $\pm$ 0.18} & 0.24 $\pm$ 0.11  & 0.15 $\pm$ 0.05 & 0.25 $\pm$ 0.04 & 0.02 $\pm$ 0.03 & \textbf{0.32 $\pm$ 0.12} \\
\textit{Gesture 5}                                             & 0.24 $\pm$ 0.20 & 0.19 $\pm$ 0.10  & 0.12 $\pm$ 0.08 & \underline{0.24 $\pm$ 0.05} & 0.22 $\pm$ 0.12 &  \textbf{0.47 $\pm$ 0.20} \\
\textit{Gesture 6}                                             & 0.21 $\pm$ 0.06 & 0.19 $\pm$ 0.05  & 0.10 $\pm$ 0.17 & 0.16 $\pm$ 0.04 & \underline{0.32 $\pm$ 0.18} & \textbf{0.40 $\pm$ 0.07}
\end{tabular}
\setcounter{table}{4}         
\renewcommand\thetable{\arabic{table}}
\caption{Gesture class-wise average macro F1 score for a motor-impaired individual with Parkinson's disease (mean$\pm$std). We report the scores after six gestures are trained with five training examples. The best macro F1 score is highlighted in bold whereas the second-best score is underlined.}
\label{tbl5}
\end{table}


\begin{table*}[ht!]
\centering
\small
\begin{tabular}{c|cccccc}
\hline
\multicolumn{1}{c|}{\multirow{2}{*}{Gesture classes}} & \multicolumn{6}{c}{\textbf{Methods}}                                          \\ \cline{2-7} 
\multicolumn{1}{c|}{}                                 & LSTM      & Vanilla-Ft & MAML-Ft   & Prototypical Net-Ft & iCaRL & \textbf{Ours-LEE} \\ \hline
\textit{Gesture 1}                                             & \underline{0.26 $\pm$ 0.14} & 0.24 $\pm$ 0.21  & 0.18 $\pm$ 0.10 & 0.18 $\pm$ 0.10 & 0.01 $\pm$ 0.01 & \textbf{0.39 $\pm$ 0.21} \\
\textit{Gesture 2}                                             & 0.13 $\pm$ 0.05 & 0.12 $\pm$ 0.03  & 0.10 $\pm$ 0.03 & \textbf{0.15 $\pm$ 0.09} & 0.01 $\pm$ 0.01 & \underline{0.14 $\pm$ 0.06} \\
\textit{Gesture 3}                                             &      
0.34 $\pm$ 0.11 & \underline{0.49 $\pm$ 0.11} & 0.22 $\pm$ 0.16 & 0.15 $\pm$ 0.08 & 0.09 $\pm$ 0.2 & \textbf{0.64 $\pm$ 0.13} \\
\textit{Gesture 4}                                             & \underline{0.29 $\pm$ 0.18} & 0.24 $\pm$ 0.10  & 0.13 $\pm$ 0.05 & 0.26 $\pm$ 0.15 & 0.06 $\pm$ 0.1 & \textbf{0.45 $\pm$ 0.21} \\
\textit{Gesture 5}                                             & 0.19 $\pm$ 0.08 & 0.14 $\pm$ 0.06  & 0.18 $\pm$ 0.06 & 0.10 $\pm$ 0.03 & \textbf{0.30 $\pm$ 0.17} &  \underline{0.30 $\pm$ 0.19} \\
\textit{Gesture 6}                                             & \underline{0.35 $\pm$ 0.17} & 0.25 $\pm$ 0.13  & 0.13 $\pm$ 0.05 & 0.22 $\pm$ 0.10 & 0.24 $\pm$ 0.19 & \textbf{0.41 $\pm$ 0.26}
\end{tabular}
\setcounter{table}{5}         
\renewcommand\thetable{\arabic{table}}
\caption{Gesture class-wise average macro F1 score for a motor-impaired individual with Multiple sclerosis (mean$\pm$std). We report the scores after six gestures are trained with five training examples. The best macro F1 score is highlighted in bold whereas the second-best score is underlined.}
\label{tbl6}
\end{table*}


\begin{table*}[ht!]
\centering
\small
\begin{tabular}{c|cccccc}
\hline
\multirow{2}{*}{Methods} & \multicolumn{6}{c}{\textit{forgetting}}                                                                                     \\ \cline{2-7} 
                         & \textit{Gesture 1} & \textit{Gesture 2} & \textit{Gesture 3} & \textit{Gesture 4} & \textit{Gesture 5} & \textit{Gesture 6} \\ \hline
LSTM                     & 0.33               & 0.49               & 0.43               & 0.12               & 0.08               & -                  \\
Vanila-Ft                & 0.47               & 0.59               & 0.29               & 0.22               & 0.11               & -                  \\
MAML-Ft                  & 0.36               & \textbf{0.39}      & 0.30               & \textbf{0.04}      & 0.04               & -                  \\
Prototypical Net-Ft      & 0.55               & 0.62               & 0.43               & 0.28               & \textbf{-0.01}              & -                  \\
iCaRL                    & 0.79                  & 0.75            & 0.53
&0.45                & -                  & -             \\
\textbf{Ours-LEE}        & \textbf{0.25}      & 0.51               & \textbf{0.30}      & 0.13               & 0.09               & -                 
\end{tabular}
\setcounter{table}{6}         
\renewcommand\thetable{\arabic{table}}
\caption{\textit{Forgetting} score in terms of gesture class-wise macro F1 score for a motor-impaired individual with Spinal cord injury in a few-shot continual learning setting (mean$\pm$std). We report the scores after six gestures are trained with five training examples. The best \textit{forgetting} score is highlighted in bold.}
\label{tbl7}
\end{table*}

\subsection{Pseudocode}
We also provide the algorithm of our LEE mechanism in Algoithm~\ref{alg:algorithm}.

\makeatletter
\setlength{\@fptop}{0pt}
\setlength{\@fpbot}{0pt plus 1fil}
\makeatother

\begin{algorithm}[ht!]
\caption{LEE for Few-Shot Continual Learning}
\label{alg:algorithm}
\textbf{Require}: {$f_\theta$}, {$f_\theta^c$}, $\alpha$, (${\mathcal{X}^{t}}, {\mathcal{Y}^{t}}$), {$\mathbf{z^c}$}
\begin{algorithmic}[1] 
\WHILE{not done}
    \STATE Compute $\mathcal{L}_{ci}$ using $\mathbf{z^c}$ and given samples from ${\mathcal{X}^{t}}$ in Equation (2)
    \STATE Compute $\mathcal{L}_{ii}$ using given samples from ${\mathcal{X}^{t}}$ in Equation (4)
    \STATE Compute $\mathcal{L}_{cls}$ using given samples and labels from ${\mathcal{X}^{t}}$ and ${\mathcal{Y}^{t}}$ in Equation (5)
    \STATE Compute $\mathcal{L}$ using $\alpha$ in Equation (1)
    \STATE Update $\theta \leftarrow \theta - \nabla_\theta\mathcal{L}$ 
\ENDWHILE
\end{algorithmic}
\end{algorithm}